\documentclass[dvipsnames]{article}

\usepackage{colm2024_conference}

\usepackage{booktabs}
\usepackage{graphicx}
\usepackage{enumitem}
\usepackage{wrapfig}
\usepackage{algorithm}
\usepackage{algpseudocode}
\usepackage{float}
\usepackage{placeins}
\usepackage{microtype}
\usepackage{amsmath}
\usepackage{colortbl}
\usepackage[utf8]{inputenc}
\usepackage{CJKutf8}
\definecolor{lightgray}{rgb}{0.9,0.9,0.9}
\usepackage{caption}
\usepackage{subcaption}
\usepackage{setspace}
\usepackage{url}
\usepackage{multirow}
\usepackage{multicol}
\usepackage{tabularx}
\usepackage{blindtext}
\usepackage{pgfplots}
\pgfplotsset{compat=1.18}
\usepackage{tikz}
\usetikzlibrary{er,positioning,bayesnet}
\usepackage{makecell}
\usepackage{tipa}
\usepackage{siunitx}
\usepackage{nicefrac}
\usepackage{tocloft}
\usepackage{listings}
\usepackage[raster,skins]{tcolorbox}
\usepackage{xltabular}
\usepackage{adjustbox}
\usepackage{xurl}
\usepackage{rotating}
\usepackage[normalem]{ulem}
\useunder{\uline}{\ul}{}
\usepackage{pifont}
\usepackage{xcolor}
\usepackage{array}
\newcommand{\checkmark}{\ding{51}}
\newcommand{\xmark}{\ding{55}}

\definecolor{qianfan_blue}{HTML}{0068B7}
\definecolor{qianfan_lightblue}{HTML}{00A0E9}
\definecolor{qianfan_gray}{HTML}{7F7F7F}

\title{Qianfan-VL: Domain-Enhanced Universal Vision-Language Models}

\author{
\bf Qianfan Team, Baidu AI Cloud
}

\begin{document}

\maketitle

\begin{abstract}
We present Qianfan-VL, a series of multimodal large language models ranging from 3B to 70B parameters, achieving state-of-the-art performance through innovative domain enhancement techniques.
Our approach employs multi-stage progressive training and high-precision data synthesis pipelines, which prove to be critical technologies for enhancing domain-specific capabilities while maintaining strong general performance.
Qianfan-VL achieves comparable results to leading open-source models on general benchmarks, with state-of-the-art performance on benchmarks such as CCBench, SEEDBench\_IMG, ScienceQA, and MMStar.
The domain enhancement strategy delivers significant advantages in OCR and document understanding, validated on both public benchmarks (OCRBench 873, DocVQA 94.75\%) and in-house evaluations.
Notably, Qianfan-VL-8B and 70B variants incorporate long chain-of-thought capabilities, demonstrating superior performance on mathematical reasoning (MathVista 78.6\%) and logical inference tasks.
All models are trained entirely on Baidu's Kunlun P800 chips, validating the capability of large-scale AI infrastructure to train SOTA-level multimodal models with over 90\% scaling efficiency on 5000+ chips for a single task. 
This work establishes an effective methodology for developing domain-enhanced multimodal models suitable for diverse enterprise deployment scenarios.
\end{abstract}

\section{Introduction}

The rapid advancement of vision-language models (VLMs) has enabled remarkable progress in multimodal understanding~\citep{radford2021learning, liu2024visual, alayrac2022flamingo}. 
However, enterprise applications often require not only general multimodal capabilities but also domain-specific expertise in critical areas such as document processing, OCR recognition, and mathematical reasoning~\citep{mathew2021docvqa, masry2022chartqa}.
Existing VLMs typically face a trade-off between maintaining broad general capabilities and achieving deep domain expertise~\citep{chen2024internvl, bai2023qwen}, despite recent advances in multimodal instruction tuning~\citep{liu2024visual} and document understanding~\citep{hu2024mplug}.

We introduce Qianfan-VL, a family of domain-enhanced vision-language models that addresses this challenge through innovative training strategies and architectural designs.
Our approach centers on three key contributions:

First, we propose a four-stage progressive training pipeline that systematically enhances domain capabilities while preserving general performance.
This pipeline progresses from cross-modal alignment (100B tokens) through general knowledge injection (2.66T tokens) and domain enhancement (0.32T tokens) to final instruction tuning (1B tokens).
The careful staging and data mixture design enables the model to acquire specialized skills without catastrophic forgetting of general knowledge.

Second, we develop comprehensive data synthesis pipelines for critical enterprise scenarios.
By combining traditional computer vision models with programmatic generation techniques, we create high-quality training data at scale.
Our synthesis covers six major task categories: document OCR, mathematical problem-solving, chart understanding, table recognition, formula recognition, and natural scene OCR.
Each pipeline incorporates domain-specific augmentation strategies and quality verification mechanisms to ensure data reliability.

Third, we demonstrate the feasibility of training large-scale VLMs entirely on proprietary hardware infrastructure.
All Qianfan-VL models are trained on Baidu's Kunlun P800 chips, utilizing innovative parallel strategies and communication-computation fusion techniques to achieve over 90\% scaling efficiency on 5000+ chip clusters.
This represents a significant milestone in developing independent AI capabilities.

Qianfan-VL offers three model variants to address different deployment scenarios:

\begin{table}[H]
    \centering
    \small
    \setlength{\tabcolsep}{8pt}
    \begin{tabular}{l c c c}
    \toprule
        \textbf{Model} & \textbf{Context} & \textbf{CoT} & \textbf{Target Deployment} \\
        \midrule
        Qianfan-VL-3B & 32K & \xmark & Edge devices, real-time OCR \\
        Qianfan-VL-8B & 32K & \checkmark & Servers, general applications \\
        Qianfan-VL-70B & 32K & \checkmark & Cloud, complex reasoning \\
    \bottomrule
    \end{tabular}
    \caption{Qianfan-VL model variants and their capabilities.}
    \label{tab:model_variants}
\end{table}

Extensive evaluations demonstrate that Qianfan-VL achieves competitive performance on general multimodal benchmarks while excelling in domain-specific tasks.
On document understanding benchmarks, Qianfan-VL-70B achieves 94.75\% on DocVQA, demonstrating strong document processing capabilities.
For mathematical reasoning, Qianfan-VL-70B reaches 78.60\% on Mathvista, demonstrating strong problem-solving capabilities.
The models also show significant improvements in OCR tasks, with scores of 873 on OCRBench~\citep{chen2024ocrbench} for the 70B variant.

\section{Model Architecture}
\label{sec:architecture}

Qianfan-VL adopts a modular architecture that combines proven components with targeted innovations for domain enhancement.
The architecture consists of three main components: a language model backbone, a visual encoder, and a cross-modal adapter.

\begin{figure}[H]
    \centering
    \includegraphics[width=\textwidth]{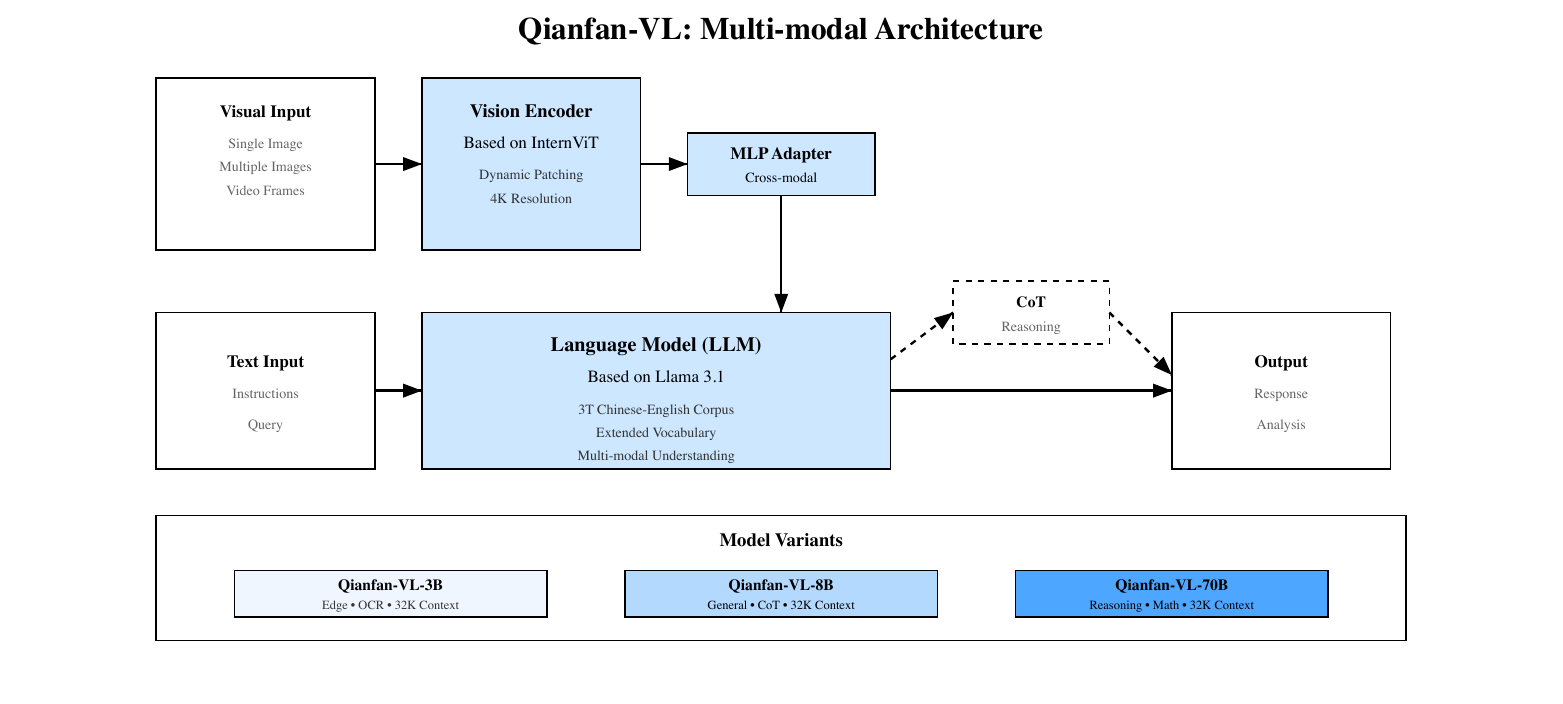}
    \caption{Overall architecture of Qianfan-VL. The model integrates the InternViT architecture for visual encoding and Llama 3.1 (8B/70B) or Qwen2.5-3B (3B) architectures for language modeling, initialized with their original pretrained weights. The cross-modal MLP adapter is randomly initialized. Based on this assembled multimodal architecture, we conduct multi-stage progressive pretraining to build the final Qianfan-VL models.}
    \label{fig:architecture}
\end{figure}

\subsection{Language Model Backbone}

The language model backbone varies across our model variants: the 8B and 70B models are based on the Llama 3.1 architecture~\citep{touvron2023llama}, while the 3B model is based on Qwen2.5-3B~\citep{bai2025qwen2}, both enhanced with vocabulary expansion and localization improvements.
We extend the original vocabulary with additional tokens and train on 3T tokens of multilingual mixed data to improve cross-lingual understanding.
The model employs Grouped-Query Attention (GQA)~\citep{ainslie2023gqa} to optimize memory efficiency and inference speed, while RMSNorm~\citep{zhang2019root} is used to improve training stability.

We provide three model variants with different parameter scales to serve diverse deployment scenarios. Table~\ref{tab:model_architecture} presents the detailed architectural parameters for each variant.

\begin{table}[H]
    \centering
    \small
    \setlength{\tabcolsep}{8pt}
    \begin{tabularx}{\textwidth}{l|>{\centering\arraybackslash}X>{\centering\arraybackslash}X>{\centering\arraybackslash}X}
    \toprule
    \textbf{Configuration} & \textbf{Qianfan-VL-3B} & \textbf{Qianfan-VL-8B} & \textbf{Qianfan-VL-70B} \\
    \midrule
    \multicolumn{4}{l}{\textit{Vision Encoder}} \\
    Hidden Size & 1024 & 1024 & 1024 \\
    \# Layers & 24 & 24 & 24 \\
    \# Num Heads & 16 & 16 & 16 \\
    Intermediate Size & 4096 & 4096 & 4096 \\
    Patch Size & 14 & 14 & 14 \\
    \midrule
    \multicolumn{4}{l}{\textit{Cross-Modal Adapter}} \\
    In Channel & 4096 & 4096 & 4096 \\
    Out Channel & 2048 & 4096 & 8192 \\
    \midrule
    \multicolumn{4}{l}{\textit{Language Model Backbone}} \\
    Hidden Size & 2048 & 4096 & 8192 \\
    \# Layers & 36 & 32 & 80 \\
    \# KV Heads & 2 & 8 & 8 \\
    Head Size & 128 & 128 & 128 \\
    Intermediate Size & 11008 & 14336 & 28672 \\
    Embedding Tying & \checkmark & $\times$ & $\times$ \\
    Vocabulary Size & 151673 & 182025 & 182025 \\
    \bottomrule
    \end{tabularx}
    \caption{Architectural parameters for Qianfan-VL model variants. Note: The 3B variant uses Qwen2.5-3B as the language model backbone, while the 8B and 70B variants use Llama 3.1.}
    \label{tab:model_architecture}
\end{table}

\subsection{Vision Encoder}

The vision encoder is initialized from InternViT~\citep{chen2024internvl} and supports dynamic image tiling for variable resolution inputs, building upon the Vision Transformer (ViT) architecture~\citep{dosovitskiy2020image}.
The encoder processes images at multiple scales, with maximum support for 4K resolution inputs.
We employ a tile-based approach where the image is divided into dynamic number of 448×448 pixel tiles, plus an additional global snapshot where the entire image is resized to 448×448 to help the model capture holistic image information.
The vision transformer configurations are as follows:

\begin{table}[H]
    \centering
    \small
    \setlength{\tabcolsep}{8pt}
    \begin{tabularx}{\textwidth}{l|>{\centering\arraybackslash}X>{\centering\arraybackslash}X>{\centering\arraybackslash}X}
    \toprule
        \textbf{Model} & \textbf{Vision Params} & \textbf{Image Tokens/Tile} & \textbf{Max Tiles} \\
        \midrule
        \textbf{Qianfan-VL-3B}  & 300M & 256 & 12 \\
        \textbf{Qianfan-VL-8B}  & 300M & 256 & 12 \\
        \textbf{Qianfan-VL-70B} & 300M & 256 & 12 \\
    \bottomrule
    \end{tabularx}
    \caption{Visual encoder configurations for Qianfan-VL models.}
    \label{tab:vision_config}
\end{table}

The dynamic tiling strategy allows the model to process high-resolution images by splitting them into multiple tiles, each processed independently and then aggregated.
This approach maintains detail preservation while managing computational costs.

\subsection{Cross-Modal Adapter}

The cross-modal adapter employs a two-layer MLP with GELU activation to project visual features into the language model's embedding space.
It starts with layer normalization on the input visual features, followed by dimensional reduction through the first linear layer, GELU activation for non-linearity, and a final linear transformation.
This design ensures stable training dynamics and efficient cross-modal alignment between vision and language representations.

\section{Training Methodology}
\label{sec:training}

\subsection{Four-Stage Progressive Training}

Our training methodology employs a carefully designed four-stage pipeline that progressively builds model capabilities:

\begin{figure}[H]
    \centering
    \includegraphics[width=\textwidth]{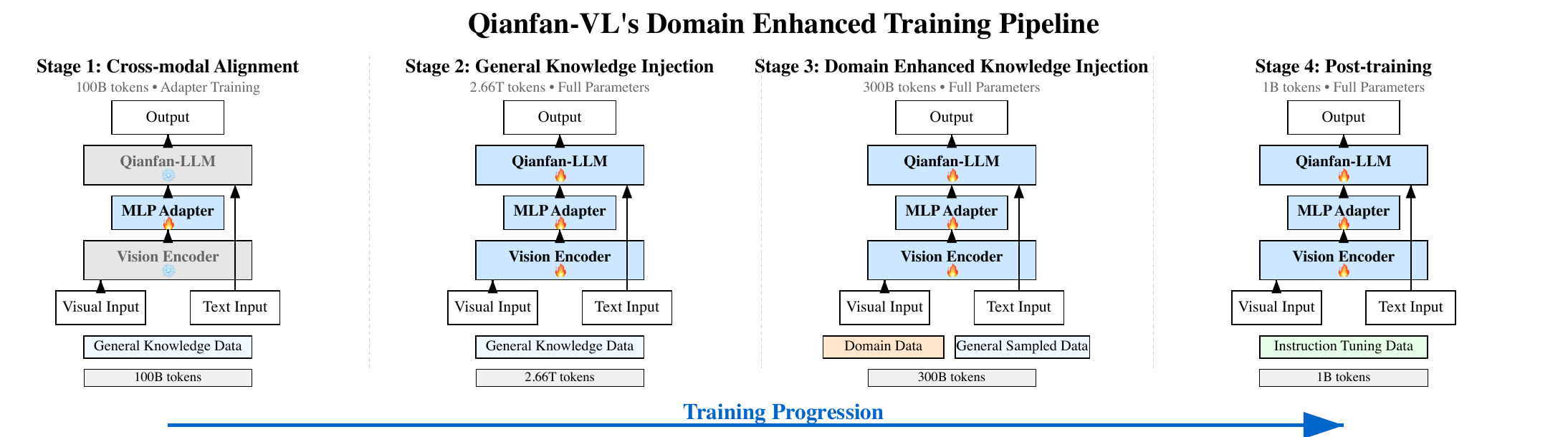}
    \caption{Four-stage progressive training pipeline of Qianfan-VL. The pipeline systematically builds capabilities from cross-modal alignment (Stage 1) through general knowledge injection (Stage 2), domain enhancement (Stage 3), to final instruction tuning (Stage 4). Each stage carefully balances different data types and training objectives to achieve domain enhancement while maintaining general capabilities.}
    \label{fig:training_pipeline}
\end{figure}

\textbf{Stage 1: Cross-Modal Alignment (100B tokens)}
In this initial stage, we establish the fundamental connection between visual and linguistic modalities to build a vision-language mapping foundation.
We adopt a conservative training strategy where only the MLP adapter parameters are updated while keeping both the vision encoder and language model completely frozen.
The training data consists of 100B tokens of high-quality image-caption pairs and basic visual question-answering tasks.
Our experiments demonstrate that this stage is necessary for stable training - without it, we observe unstable loss curves during the early phase of Stage 2, which negatively impacts final model performance.
The frozen encoder strategy ensures that pre-trained representations remain intact while the adapter learns to bridge the modality gap.

\textbf{Stage 2: General Knowledge Injection (2.66T tokens)}
This stage focuses on injecting massive amounts of general knowledge while establishing robust multimodal understanding through our comprehensive dataset collection.
We perform full parameter updates across all model components - vision encoder, language model, and adapter.

\begin{figure}[H]
    \centering
    \includegraphics[width=0.9\textwidth]{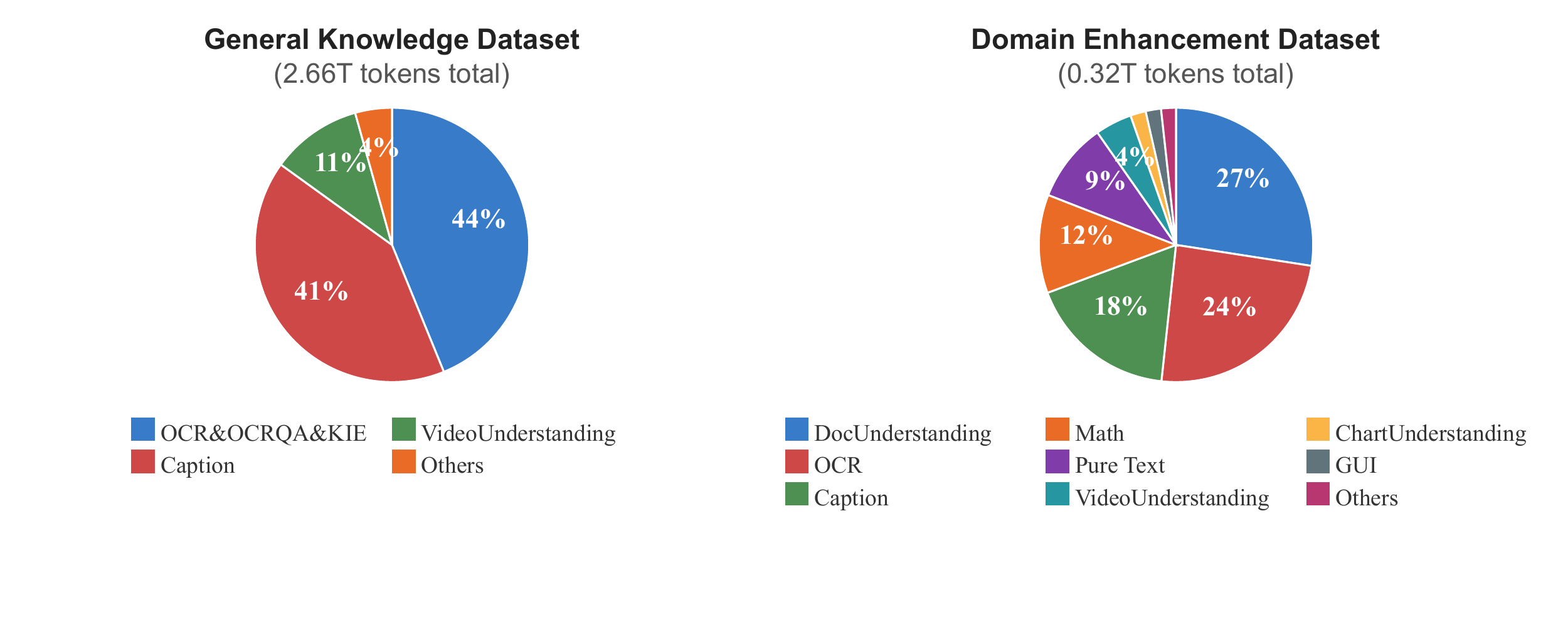}
    \caption{Distribution of training data across different task categories. Left: General Knowledge Dataset (2.66T tokens total) with OCR\&OCRQA\&KIE (43.8\%) and Caption (41.1\%) tasks dominating, VideoUnderstanding (10.7\%), and Others (4.3\%) including Grounding, ChartUnderstanding, DocUnderstanding, GUI, Knowledge, Math, and VQA. Right: Domain Enhancement Dataset (0.32T tokens total) with more balanced distribution across specialized domains.}
    \label{fig:ocean_all_distribution}
\end{figure}

The General Knowledge Dataset comprises 2.66T tokens distributed across major task categories, as shown in Figure~\ref{fig:ocean_all_distribution}. The dataset emphasizes document understanding and visual description capabilities, with OCR\&OCRQA\&KIE tasks (43.8\%) combining open-source datasets including EATEN~\citep{guo2019eaten}, VCR~\citep{zhang2024vcr}, CASIA~\citep{liu2020offline}, WKVVQA~\citep{shah2019kvqa}, LLaVAR~\citep{zhang2023llavar}, and A-OKVQA~\citep{schwenk2022aokvqa} with proprietary OCR synthesis datasets for robust text recognition and key information extraction. Caption generation (41.1\%) leverages ShareGPT4Video~\citep{chen2024sharegpt4video}, GRIT~\citep{grit2023}, MedTrinity~\citep{xie2024medtrinity25mlargescalemultimodaldataset}, WebSight~\citep{laurencon2024unlockingconversionwebscreenshots}, and COYO-700M along with in-house video caption datasets. VideoUnderstanding (10.7\%) integrates LLaVA-Video-178K and ChinaOpen with proprietary video datasets for temporal reasoning. The remaining tasks (4.3\%) encompass Grounding (All-Seeing-V2, V3Det, RefCOCO/+/g~\citep{kazemzadeh2014referitgame}), ChartUnderstanding~\citep{masry2022chartqa}, DocUnderstanding~\citep{mathew2021docvqa}, GUI understanding, general Knowledge tasks, Math reasoning datasets, and VQA benchmarks, ensuring comprehensive coverage of multimodal understanding scenarios.

\textbf{Stage 3: Domain Enhancement (0.32T tokens)}
This critical stage implements targeted capability enhancement for enterprise-critical domains.
We maintain full parameter updates while carefully balancing the training mixture with 70\% meticulously curated domain-specific data and 30\% general data sampling to maintain broad capabilities.
As shown in the right panel of Figure~\ref{fig:ocean_all_distribution}, the Domain Enhancement Dataset (0.32T tokens) has a more balanced distribution across specialized domains. DocUnderstanding (27.4\%) focuses on comprehensive document analysis including contracts, invoices, reports, and academic papers with complex layout understanding. OCR (24.3\%) addresses advanced text recognition tasks including handwriting, scene text, formula recognition, and multi-language scripts. Caption generation (17.6\%) targets domain-specific visual descriptions for specialized content and technical imagery. Mathematical reasoning (11.6\%) covers K-12 to university-level problems with detailed solutions and proof verification. Pure Text Pretraining Data (9.4\%) maintains language modeling capabilities, while VideoUnderstanding (4.3\%) provides specialized video analysis, ChartUnderstanding (1.8\%) enables business intelligence and data visualization interpretation, and GUI tasks (1.8\%) support user interface understanding. We employ curriculum learning with adaptive difficulty scheduling, starting with simple OCR tasks and progressively introducing complex multi-step reasoning problems, ensuring stable learning and preventing overfitting to specific task patterns.

\textbf{Stage 4: Post-training with Instruction Tuning (1B tokens)}
The final stage focuses on post-training through comprehensive instruction tuning to enhance the model's instruction following capabilities.
We continue full parameter updates with a carefully curated dataset of 1B tokens encompassing complex instruction following (multi-step tasks, conditional logic, edge cases), writing and generation (reports, summaries, creative content), question answering (factual, analytical, and reasoning-based queries), programming assistance (code understanding, debugging, documentation), and domain-specific instructions (OCR formatting, mathematical notation, chart interpretation).
For tasks requiring logical reasoning and mathematical computation—such as chart question answering, visual problem solving, and visual reasoning—we employ long chain-of-thought (Long CoT) techniques to significantly enhance the model's reasoning capabilities.
These Long CoT traces provide detailed step-by-step reasoning paths and intermediate computational steps, enabling the model to tackle complex multi-step problems with improved accuracy.
Critically, we include substantial pure-text instruction data to maintain the language model's capabilities while systematically improving its ability to handle reasoning-intensive visual tasks.
Additionally, we perform model merging on the best-performing checkpoints from different training runs to combine their complementary strengths, resulting in enhanced overall performance across all evaluation metrics.

\subsection{Data Synthesis Pipeline}

\begin{figure}[!htbp]
\centering
\includegraphics[width=\textwidth]{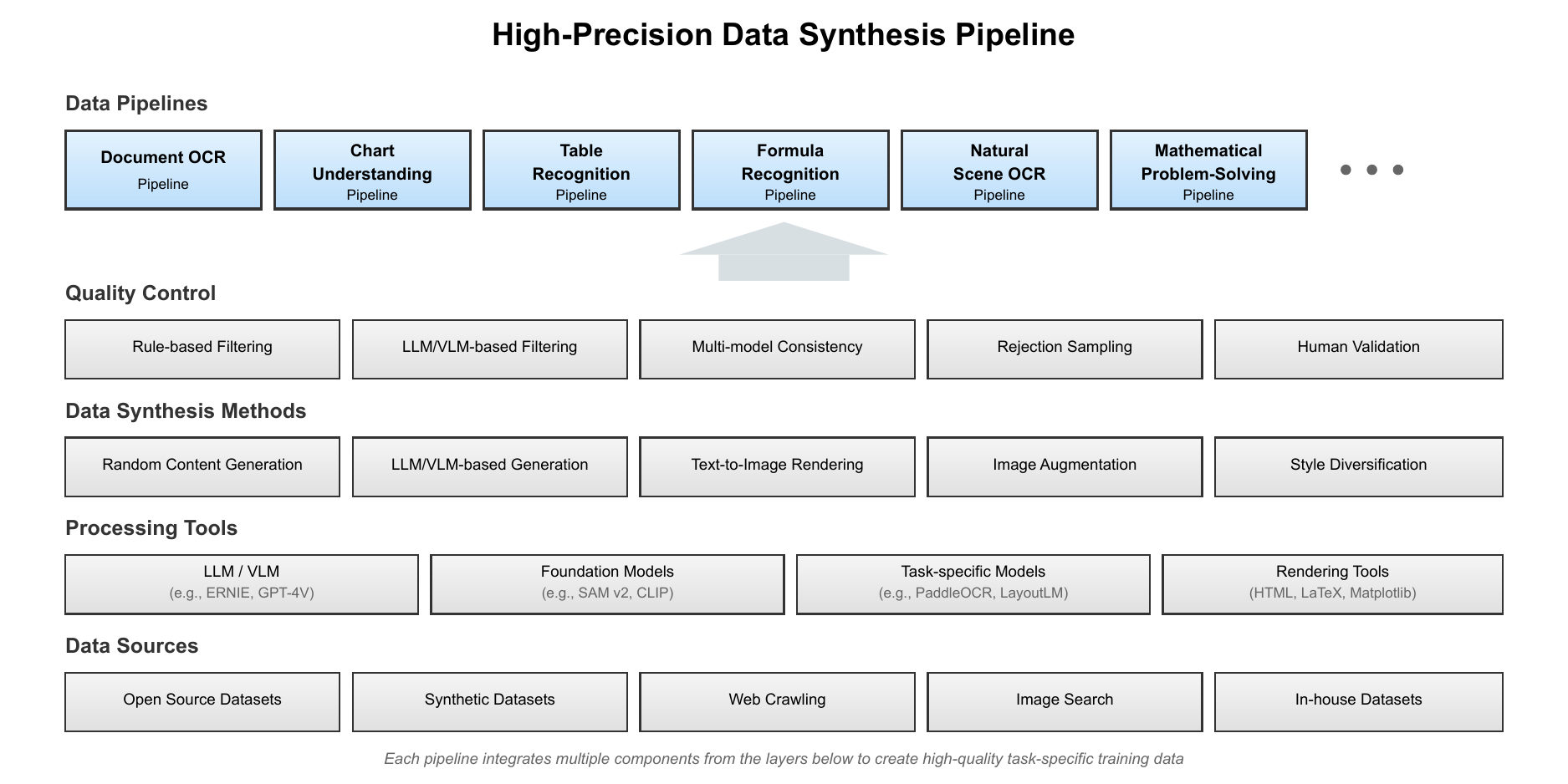}
\caption{Overview of the data synthesis pipeline architecture, showing the integrated workflow from data sources through various processing stages including synthesis tools, quality control, and specialized data pipelines for different modalities.}
\label{fig:data_synthesis_pipeline}
\end{figure}

We develop comprehensive data synthesis pipelines for six major task categories, combining traditional computer vision models with programmatic generation to create high-quality training data at unprecedented scale.
Our synthesis approach emphasizes diversity, accuracy, and real-world applicability:

\begin{table}[H]
    \centering
    \small
    \setlength{\tabcolsep}{8pt}
    \begin{tabular}{l l}
    \toprule
        \textbf{Pipeline} & \textbf{Key Features} \\
        \midrule
        Document OCR & Multi-format, noise simulation \\
        Mathematics & K-12 to university, step-by-step \\
        Charts & 15+ types, Q\&A pairs \\
        Tables & Complex structures, 50+ themes \\
        Formulas & Multi-engine, handwriting \\
        Scene OCR & Natural embedding, multilingual \\
    \bottomrule
    \end{tabular}
    \caption{Data synthesis pipeline summary showing key characteristics of each category.}
    \label{tab:data_synthesis}
\end{table}

\textbf{Document OCR Pipeline:}
Our document OCR pipeline implements three core functionalities: full document parsing through multi-dimensional analysis combining layout detection, content extraction, and structural understanding (supporting multilingual documents including handwritten and scanned materials); image-to-Markdown conversion for efficient transformation of single/multi-page documents into structured format preserving formatting and hierarchy; and document Q\&A capabilities supporting summarization, reasoning, and multi-turn dialogue about content.
Data sources include DocVQA~\citep{mathew2021docvqa}, DocReason25K, and proprietary synthesized datasets, with robustness enhancements applied through bitmap rendering, morphological operations (erosion/dilation), Gaussian blur, and other noise simulation techniques inspired by document augmentation approaches~\citep{groleau2022augraphy}.
Quality assurance employs multi-VLM cross-validation with agreement thresholds, and the pipeline generates documents with varying complexity levels, from simple forms to complex multi-column layouts with embedded tables and figures.

\textbf{Mathematical Problem-Solving Pipeline:}
Our mathematical pipeline addresses the unique challenges of educational scenarios through comprehensive educational data preprocessing that collects multilingual high-quality problem-solving data with standardized terminology and notation, performing structured decomposition into problem statements, conditions, solution steps, and formulas.
The solution synthesis follows a knowledge-guided generation pipeline from structured representation through LaTeX formatting and HTML rendering to image generation, creating realistic photo-solving scenarios with specialized handling of charts, formulas, and geometric figures using Markdown, LaTeX, and Asymptote formal description languages.
We enhance diversity through multiple handwriting styles, paper backgrounds, and lighting conditions to simulate K-12 through university-level scenarios, while quality verification employs rule-based filtering, rejection sampling, multi-model voting, and OCR character-level validation to ensure mathematical correctness.
The pipeline covers algebra, geometry, calculus, statistics, and linear algebra with adaptive difficulty scaling.

\textbf{Chart Understanding Pipeline:}
Our chart understanding pipeline automates the generation of high-quality chart Q\&A pairs through a comprehensive process beginning with data expansion via open-source dataset sampling combined with web crawling through image search APIs, followed by deduplication.
Pre-trained VLMs generate structured summaries containing both visual and numerical information, which feed into a two-stage generation process where questions are first generated from summaries, then answers are produced based on questions and summaries.
We incorporate LaTeX rendering through ArXiv paper crawling with regex extraction and TexLive re-rendering for precise mathematical chart descriptions inspired by Nougat~\citep{blecher2023nougat}, while quality control employs chain-of-thought model verification combined with human review to ensure accuracy.
The pipeline produces three question types: data retrieval (exact value extraction), visual attributes (color, style, layout), and computational Q\&A (aggregation, comparison, trend analysis), covering 15+ chart types including bar, line, pie, scatter, heatmap, box plot, and complex composite visualizations.

\textbf{Table Recognition Pipeline:}
Our table pipeline addresses two core capabilities: table structure recovery for precise conversion of image tables to HTML/LaTeX (supporting borderless tables, contract forms, and complex layouts with merged cells), and table Q\&A for numerical computation, comparative analysis, and information retrieval based on table images.
The synthesis process employs content generation through random table structures (3-20 rows/columns) populated via the Faker library and LLM-based realistic data filling with random cell merging, combined with visual rendering using 50+ professional CSS themes (statistical reports, technical documents, financial statements) rendered through Jinja2+KaTeX engines.
Data augmentation applies geometric transformations, color perturbations, and blur effects for diversity, drawing from sources including TabMWP~\citep{lu2022dynamic}, MMC-Inst~\citep{liu2023mmc}, BigDocs~\citep{rodriguez2024bigdocs}, and proprietary synthetic datasets.
The pipeline handles complex scenarios like multi-level headers, footnotes, and cross-references.

\textbf{Formula Recognition Pipeline:}
Our formula pipeline achieves symbol recognition, syntax parsing, and semantic understanding through comprehensive symbol coverage (mathematical symbols, Greek letters, special notations, and domain-specific markers) and structure parsing of complex elements including fractions, radicals, subscripts/superscripts, matrices, and tensor notations.
Semantic mapping links formula semantics to mathematical concepts for deeper understanding, while multi-engine rendering with MathJax and KaTeX ensures cross-platform consistency, complemented by handwriting simulation featuring diverse writing styles, paper textures, ink variations, and noise patterns.
The dataset spans from elementary topics (arithmetic operations, basic algebra) through intermediate concepts (trigonometry, logarithms, sequences) to advanced mathematics (calculus, differential equations, linear algebra, abstract algebra), with each formula including LaTeX source, rendered image, and semantic annotations.

\textbf{Natural Scene OCR Pipeline:}
Our pipeline, inspired by SynthText~\citep{gupta2016synthetic}, implements systematic text-in-image synthesis through background screening using lightweight OCR models~\citep{cui2025paddleocr} and image type detection to eliminate text-containing and non-static samples, combined with scene understanding via semantic segmentation~\citep{ravi2024sam} and monocular depth estimation~\citep{yang2024depth} for region division and 3D structure.
Realistic projection employs plane detection with perspective transformation and random text styling for natural embedding, while fusion enhancement through Poisson blending ensures consistent occlusion, shadows, and texture integration.
The pipeline generates diverse scenes including street environments (signs, storefronts, billboards), indoor settings (product labels, menus, notices), and documents in context (whiteboards, presentations, posters), with annotations providing character-level and word-level bounding boxes along with reading order information.
Text languages cover over 12 languages with appropriate fonts and styles.

\subsection{Complex Instruction Enhancement}

Enterprise applications require sophisticated instruction-following beyond simple queries.
We address this gap by evolving simple prompts into multi-constraint instructions through a systematic pipeline:

\textbf{(1) Seed Construction and Mining:}
We select domain-relevant images paired with initial prompts, then expand coverage using SFTMiner, our retrieval engine supporting text-to-image and image-to-image search across massive repositories.

\textbf{(2) Prompt Evolution:}
Starting from 5 manually designed seed prompts per scenario, we generate 5-10 simple single-constraint variants, then systematically introduce diverse constraint types (conditional reasoning, quantity limitations, sequential constraints) to create complex multi-step instructions.

\textbf{(3) Response Generation:}
We generate responses using advanced VLMs, forming \texttt{<image, prompt, response>} triplets with multi-model voting and consistency checking for quality assurance.

Through this pipeline, we synthesize approximately 200K complex instruction samples covering conditional extraction, multi-constraint analysis, reasoning with rejection handling, and hierarchical decomposition, significantly enhancing the model's ability to handle sophisticated enterprise requirements.

\subsection{Chain-of-Thought Training}

\begin{figure}[H]
    \centering
    \includegraphics[width=0.95\textwidth]{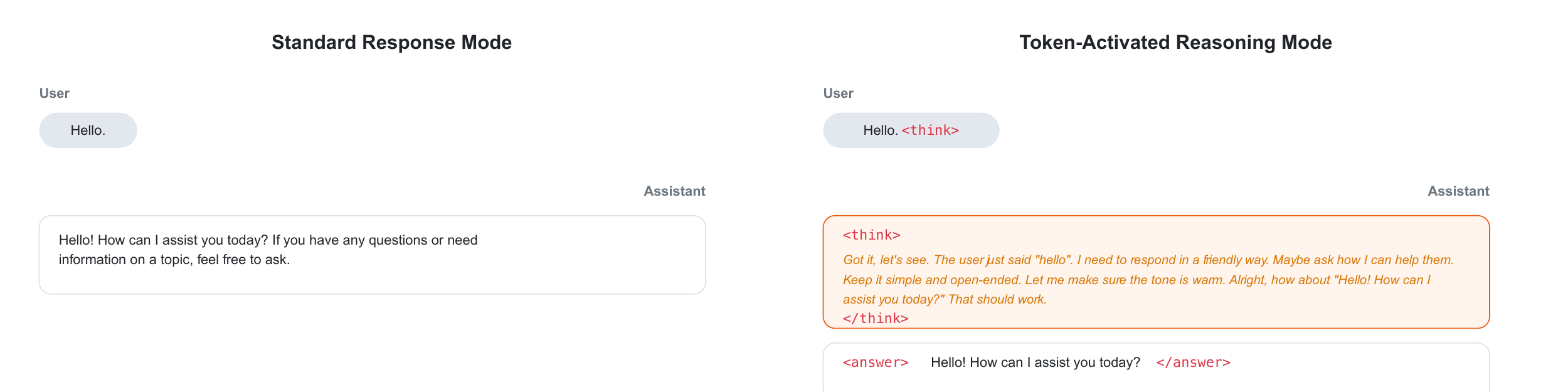}
    \caption{Comparison between standard response mode and token-activated reasoning mode. The left panel shows direct response generation, while the right panel demonstrates how \texttt{<think>} tokens trigger internal reasoning processes (shown in orange dashed box) that are hidden from the user, with only the final answer visible.}
    \label{fig:reasoning_mode}
\end{figure}

For the 8B and 70B variants, we implement sophisticated chain-of-thought (CoT) reasoning capabilities~\citep{wei2022chain} through a multi-faceted approach:

\textbf{Token-Activated Reasoning:}
We introduce special tokens (\texttt{<think>} and \texttt{</think>}) to delineate reasoning processes, allowing users to explicitly request reasoning by including these tokens.
The model learns to generate intermediate reasoning steps within these boundaries while providing final answers outside the thinking tokens for clarity.
Figure~\ref{fig:reasoning_mode} illustrates the difference between standard mode and token-activated reasoning mode, showing how the thinking tokens enable explicit chain-of-thought generation.

\textbf{Training Data Construction:}
Our CoT training corpus primarily consists of approximately 200K mathematical reasoning problems, many of which are synthesized into multimodal data by combining mathematical problems with visual elements.
We leverage advanced models with thinking capabilities such as DeepSeek-R1 to generate long chain-of-thought data with detailed step-by-step reasoning processes, ensuring comprehensive coverage of mathematical concepts from basic arithmetic to advanced calculus.

\textbf{Quality Assurance:}
We employ rejection sampling to generate 5-10 solutions per problem and select the best via reward models, combined with process supervision to verify correctness of intermediate steps rather than just final answers.
Consistency checking ensures reasoning chains are logically coherent and complete, while human validation through expert review of 10\% of samples maintains quality standards.

\section{Infrastructure and Implementation}
\label{sec:infrastructure}

\subsection{Kunlun Chip Training}

All Qianfan-VL models are trained on Baidu's Kunlun P800 chips, demonstrating the viability of proprietary AI infrastructure for large-scale model development.

\textbf{Model Adaptation and Selection:}
Our infrastructure successfully adapted multiple foundation model families to the Kunlun P800 architecture, including the Llama 3.1 series across various scales and the Qwen2.5 series in different parameter configurations.
We also successfully adapted the InternViT vision encoder for efficient visual processing on Kunlun chips.
Through extensive ablation experiments evaluating different model combinations and scales, we determined the optimal configurations: 70B and 8B models based on Llama 3.1 architecture for their superior performance on general multimodal tasks, and the 3B model based on Qwen2.5 for its excellent efficiency in edge deployment scenarios.
These experiments fully validate the capability of both Llama and Qwen series models to undergo parallel training on large-scale Kunlun chip clusters with massive datasets, achieving excellent performance outcomes.
This demonstrates the maturity of proprietary AI infrastructure in supporting diverse model architectures and training paradigms at scale.

\textbf{Cluster Configuration:}
Our training infrastructure comprises over 5000 Kunlun P800 chips operating in parallel, processing a massive 3T token training corpus while maintaining over 90\% scaling efficiency at large scale.

\textbf{3D Parallelism Strategy:}
Our parallelism strategy combines three dimensions for optimal scaling: Data Parallelism (DP) distributes batch samples across nodes with gradient accumulation, Tensor Parallelism (TP) splits model layers across chips within nodes for memory efficiency, and Pipeline Parallelism (PP) divides model depth across node groups to maximize throughput.

Key optimizations include dynamic load balancing with adaptive distribution based on layer computation patterns, optimized AllReduce gradient synchronization achieving 60\% communication reduction, and 1F1B (one forward, one backward) pipeline scheduling with bubble rate below 5\%.
Additionally, sequence parallelism for splitting long sequences reduces memory by 50\% for 32K contexts, while dynamic batching adapts batch sizes based on sequence length distribution, and selective recomputation strategically places checkpoints to balance memory and computation.

\textbf{Communication-Computation Fusion:}
\textit{Architectural Advantage:}
The Kunlun P800 chip features a unique architecture where communication and matrix multiplication units are physically separated, unlike traditional GPUs where they compete for resources.
This hardware design enables resource isolation where communication operations never block computation resources, true parallelism with simultaneous data transfer and matrix operations, and overlap optimization that hides communication latency behind computation.

\textit{GEMM Fusion Implementation:}
We establish bypass streams (BypassStream) for seamless integration, enabling independent scheduling where bypass streams run parallel to main computation streams, data prefetching that initiates communication before computation needs data, and result pipelining for immediate transfer of computation results.

\textit{Multi-Stream Optimization:}
Taking AllGather+GEMM fusion as an example, traditional approaches complete AllGather, wait, then start GEMM sequentially.
Our optimized approach transfers and computes data blocks in a pipeline fashion, achieving a 40\% reduction in end-to-end latency for large operations.

\section{Evaluation Results}
\label{sec:evaluation}

We conduct comprehensive evaluations across general multimodal benchmarks and domain-specific tasks using the VLMEvalKit framework~\citep{duan2024vlmevalkit}, an open-source toolkit designed for evaluating large multi-modality models.
Most benchmarks employ the framework's built-in rule-based evaluation methods for consistent and reproducible assessment.
For benchmarks requiring more nuanced evaluation—including mathematical reasoning tasks, subjective assessments, and those with strict output format requirements—we implement LLM-as-a-judge evaluation using Ernie-4.5-Turbo-VL~\citep{ernie2025technicalreport}, achieving over 95\% accuracy as validated through manual assessment.

\subsection{General Multimodal Benchmarks}

We conduct extensive evaluations across 14 standard multimodal benchmarks to assess general visual understanding capabilities.
Table~\ref{tab:general_benchmarks} presents comprehensive comparisons with state-of-the-art open-source models including InternVL 3~\citep{zhu2025internvl3} and Qwen2.5-VL~\citep{bai2025qwen2}.

\begin{table*}[!htbp]
\centering
\small
\setlength{\tabcolsep}{7pt}
\begin{tabular}{l|ccc|cc|cc}
\toprule
 & \multicolumn{3}{c|}{Qianfan-VL} & \multicolumn{2}{c|}{InternVL-3} & \multicolumn{2}{c}{Qwen2.5-VL} \\
Benchmark & 3B & 8B & 70B & 8B & 78B & 7B & 72B \\
\midrule
A-Bench\_VAL~\citep{zhang2024bench} & 75.65 & 75.72 & \textbf{78.10} & 75.86 & 75.86 & 76.49 & \textbf{79.22} \\
CCBench~\citep{liu2024mmbench} & 66.86 & 70.39 & \textbf{80.98} & 77.84 & 70.78 & 57.65 & \textbf{73.73} \\
SEEDBench\_IMG~\citep{li2023seed} & 76.55 & 78.02 & \textbf{79.13} & 77.0 & 77.52 & 76.98 & \textbf{78.34} \\
SEEDBench2\_Plus~\citep{li2024seed} & 67.59 & 70.97 & \textbf{73.17} & 69.52 & 68.47 & 70.93 & \textbf{73.25} \\
MMVet~\citep{yu2023mm} & 48.17 & 53.21 & 57.34 & \textbf{80.28} & \textbf{78.9} & 70.64 & 75.69 \\
MMMU\_VAL~\citep{yue2023mmmu} & 46.44 & 47.11 & 58.33 & 56.11 & \textbf{60.78} & 51.0 & \textbf{65.78} \\
ScienceQA\_TEST~\citep{lu2022learn} & 95.19 & 97.62 & \textbf{98.76} & 97.97 & 97.17 & 85.47 & 92.51 \\
ScienceQA\_VAL~\citep{lu2022learn} & 93.85 & 97.62 & \textbf{98.81} & \textbf{97.81} & 95.14 & 83.59 & 91.32 \\
MMT-Bench\_VAL~\citep{ying2024mmt} & 62.23 & 63.22 & \textbf{71.06} & 65.17 & 63.67 & 61.4 & \textbf{69.49} \\
MTVQA\_TEST~\citep{tang2024mtvqa} & 26.50 & 30.14 & \textbf{32.18} & 30.30 & 27.62 & 29.08 & \textbf{31.48} \\
BLINK~\citep{fu2024blink} & 49.97 & 56.81 & \textbf{59.44} & 55.87 & 51.87 & 54.55 & \textbf{63.02} \\
MMStar~\citep{chen2024we} & 57.93 & 64.07 & \textbf{69.47} & \textbf{68.40} & 66.07 & 61.53 & 66.00 \\
RealWorldQA~\citep{ai2024grok} & 65.75 & 70.59 & 71.63 & 71.11 & \textbf{74.25} & 69.28 & \textbf{73.86} \\
Q-Bench1\_VAL~\citep{wu2023q} & 73.51 & 75.25 & 77.46 & 75.99 & 77.99 & \textbf{78.10} & \textbf{79.93} \\
POPE~\citep{li2023evaluating} & 85.08 & 86.06 & 88.97 & \textbf{90.59} & 88.87 & 85.97 & 83.35 \\
RefCOCO (Avg)~\citep{kazemzadeh2014referitgame} & 85.94 & 89.37 & \textbf{91.01} & 89.65 & \textbf{91.40} & 86.56 & 90.25 \\
\bottomrule
\end{tabular}
\caption{Performance on general multimodal benchmarks. Top-2 results for each benchmark are in bold.}
\label{tab:general_benchmarks}
\end{table*}

Qianfan-VL demonstrates competitive performance across general benchmarks:
\begin{itemize}[noitemsep,topsep=0pt]
\item \textbf{Scientific reasoning:} Achieves 98.17\% on ScienceQA\_TEST (8B/70B), surpassing most comparable models
\item \textbf{Hallucination resistance:} Strong performance on POPE (88.79\% for 70B) indicates robust grounding
\item \textbf{Visual perception:} Competitive scores on SEEDBench\_IMG (78.85\% for 70B) and Q-Bench (78.33\% for 70B)
\item \textbf{Real-world understanding:} Solid performance on RealWorldQA (71.76\% for 70B) demonstrates practical applicability
\end{itemize}

However, we observe relatively lower performance on MMMU (46.44\%-58.33\%) and MMVet (48.17\%-57.34\%) compared to leading models.
Analysis of failure cases reveals that these gaps primarily stem from insufficient coverage of general knowledge questions, particularly those requiring broad understanding of diverse topics and complex reasoning across multiple domains.
This limitation can be addressed in future iterations by incorporating more interleaved image-text data covering general knowledge domains, which would enhance the model's ability to handle open-ended questions requiring extensive world knowledge while maintaining our strong domain-specific capabilities.

Despite these areas for improvement, Qianfan-VL achieves competitive results across most benchmarks while being optimized for domain-specific tasks, validating our approach of enhancing specialized capabilities without sacrificing overall general performance.

\subsection{OCR and Document Understanding}

Table~\ref{tab:ocr_benchmarks} shows performance on OCR and document understanding tasks, where Qianfan-VL exhibits significant advantages.

\begin{table*}[!htbp]
\centering
\small
\setlength{\tabcolsep}{6pt}
\begin{tabular}{l|ccc|cc|ccc}
\toprule
 & \multicolumn{3}{c|}{Qianfan-VL} & \multicolumn{2}{c|}{InternVL-3} & \multicolumn{3}{c}{Qwen2.5-VL} \\
Benchmark & 3B & 8B & 70B & 8B & 78B & 3B & 7B & 72B \\
\midrule
OCRBench~\citep{chen2024ocrbench} & 831 & 854 & 873 & \textbf{881} & 847 & 810 & \textbf{883} & 874 \\
AI2D\_TEST~\citep{kembhavi2016diagram} & 81.38 & \textbf{85.07} & \textbf{87.73} & 85.07 & 83.55 & 77.07 & 80.47 & 83.84 \\
OCRVQA\_TEST~\citep{mishra2019ocr} & 66.15 & 68.98 & \textbf{74.06} & 39.03 & 35.58 & 69.24 & \textbf{71.02} & 66.8 \\
TextVQA\_VAL~\citep{singh2019towards} & 80.11 & 82.13 & \textbf{84.48} & 82.15 & 83.52 & 79.09 & \textbf{84.96} & 83.26 \\
DocVQA\_VAL~\citep{mathew2021docvqa} & 90.85 & 93.54 & \textbf{94.75} & 92.04 & 83.82 & 92.71 & \textbf{94.91} & \textbf{95.75} \\
ChartQA\_TEST~\citep{masry2022chartqa} & 81.79 & \textbf{87.72} & \textbf{89.60} & 85.76 & 82.04 & 83.40 & 86.68 & 87.16 \\
\bottomrule
\end{tabular}
\caption{Performance on OCR and document understanding benchmarks. Top-2 results for each benchmark are in bold.}
\label{tab:ocr_benchmarks}
\end{table*}

The domain enhancement strategy yields exceptional results:
\begin{itemize}[noitemsep,topsep=0pt]
\item \textbf{Document understanding:} Qianfan-VL-70B achieves 94.75\% on DocVQA, with the series maintaining competitive performance across all model sizes
\item \textbf{Chart analysis:} 89.60\% on ChartQA\_TEST (70B), leading among comparable model sizes
\item \textbf{OCR accuracy:} Strong showing on OCRBench~\citep{chen2024ocrbench} (873 for 70B), with competitive performance across the model series
\item \textbf{Visual question answering:} Competitive on TextVQA (84.48\% for 70B) and OCRVQA (74.06\% for 70B)
\end{itemize}
These results validate our targeted data synthesis and domain-focused training approach, showing particular strength in document-centric tasks critical for enterprise applications.

\subsection{Mathematical Reasoning}

Mathematical reasoning capabilities are evaluated on specialized benchmarks as shown in Table~\ref{tab:math_benchmarks}.

\begin{table*}[!htbp]
\centering
\small
\setlength{\tabcolsep}{10pt}
\begin{tabular}{l|cc|cc|cc}
\toprule
 & \multicolumn{2}{c|}{Qianfan-VL} & \multicolumn{2}{c|}{InternVL-3} & \multicolumn{2}{c}{Qwen2.5-VL} \\
Benchmark & 8B & 70B & 8B & 78B & 7B & 72B \\
\midrule
Mathvista-mini~\citep{lu2023mathvista} & 69.19 & \textbf{78.60} & 69.50 & 70.10 & 67.20 & 73.90 \\
Mathvision~\citep{wang2024measuring} & 32.82 & \textbf{50.29} & 29.61 & 34.8 & 25.95 & 39.34 \\
Mathverse~\citep{zhang2024mathverse} & 48.4 & \textbf{61.04} & 43.68 & 49.26 & 44.21 & 55.18 \\
ChartQA Pro~\citep{masry2025chartqapro} & 50.41 & \textbf{52.00} & 37.32 & 44.43 & 43.73 & 45.30 \\
HallusionBench~\citep{guan2024hallusionbench} & 51.72 & \textbf{54.52} & 49.20 & 40.20 & 47.90 & 49.90 \\
InHouse Dataset A & 59.87 & \textbf{71.78} & 40.64 & 41.47 & 45.58 & 57.20 \\
InHouse Dataset B & 61.33 & \textbf{75.60} & 36.25 & 42.65 & 30.62 & 59.68 \\
\bottomrule
\end{tabular}
\caption{Mathematical reasoning performance with CoT enabled. Best results for each benchmark are in bold.}
\label{tab:math_benchmarks}
\end{table*}

The chain-of-thought training delivers remarkable improvements in mathematical and visual reasoning:
\begin{itemize}[noitemsep,topsep=0pt]
\item \textbf{MathVista performance:} Qianfan-VL-70B achieves 78.60\% on Mathvista-mini, demonstrating state-of-the-art performance among open-source models
\item \textbf{Complex visual math:} 50.29\% on Mathvision (70B), significantly outperforming comparable open models
\item \textbf{Multi-step reasoning:} 61.04\% on Mathverse (70B), demonstrating strong capability in problems requiring multiple reasoning steps
\item \textbf{Chart-based reasoning:} 52.00\% on ChartQA Pro (70B), leading among open-source models
\item \textbf{Hallucination mitigation:} 54.52\% on HallusionBench (70B), showing improved factual grounding with CoT
\item \textbf{Proprietary benchmarks:} Strong performance on internal datasets (71.78\% and 75.60\% for 70B) validates real-world applicability
\end{itemize}
The consistent improvements across diverse reasoning tasks demonstrate that our CoT training methodology effectively enhances complex problem-solving capabilities beyond simple pattern matching.

\subsection{Ablation Studies On The Validity of Domain Enhancement}

We conduct comprehensive ablation studies to validate the effectiveness of our domain enhancement strategy (Stage 3). To isolate the impact of domain-specific training, we fix the data in Stages 1, 2, and 4 while comparing models trained with and without Stage 3.

\begin{table*}[!htbp]
    \centering
    \small
    \setlength{\tabcolsep}{5pt}
    \begin{tabular}{l|ccccc}
    \toprule
    \multirow{2}{*}{\textbf{Config}} & \multicolumn{5}{c}{\textbf{Document Understanding Tasks}} \\
    & DocVQA & AI2D & ChartQA & Doc Text* & Simple HTML* \\
    \midrule
    w/o Stage 3 & 93.67 & 84.59 & 86.68 & 73.70 & 68.70 \\
    w/ Stage 3 & 94.13 & 85.33 & 88.00 & 75.00 & 72.10 \\
    \midrule
    \textbf{Gain} & +0.46 & +0.74 & +1.32 & +1.30 & +3.40 \\
    \bottomrule
    \end{tabular}
    \hfill
    \begin{tabular}{l|cccccc}
    \toprule
    \multirow{2}{*}{\textbf{Config}} & \multicolumn{6}{c}{\textbf{OCR Tasks}} \\
    & OCRBench & Complex HTML* & Chart Dict* & LaTeX* & Struct Extract* & Handwriting* \\
    \midrule
    w/o Stage 3 & 837 & 78.33 & 75.73 & 85.04 & 33.17 & 75.69 \\
    w/ Stage 3 & 852 & 82.00 & 77.21 & 86.16 & 35.85 & 83.89 \\
    \midrule
    \textbf{Gain} & +15 & +3.67 & +1.48 & +1.12 & +2.68 & +8.20 \\
    \bottomrule
    \end{tabular}
    \caption{Impact of Stage 3 (Domain Enhancement) on OCR and document understanding tasks. * indicates in-house datasets.}
    \label{tab:ablation_ocr}
\end{table*}

\begin{table}[H]
    \centering
    \small
    \setlength{\tabcolsep}{6pt}
    \begin{tabular}{l|ccccc}
    \toprule
    \multirow{2}{*}{\textbf{Configuration}} & \multicolumn{5}{c}{\textbf{Mathematical Benchmarks}} \\
    & Mathvista-mini & Mathvision & Mathverse & In-House A & In-House B \\
    \midrule
    w/o Stage 3 & 75.50 & 45.52 & 57.34 & 59.28 & 65.83 \\
    w/ Stage 3 & 77.10 & 51.54 & 59.95 & 77.28 & 73.01 \\
    \midrule
    \textbf{Gain} & +1.60 & +6.02 & +2.61 & +18.00 & +7.18 \\
    \bottomrule
    \end{tabular}
    \caption{Impact of Stage 3 (Domain Enhancement) on mathematical reasoning tasks.}
    \label{tab:ablation_math}
\end{table}

\textbf{Key Findings:}
The ablation results clearly demonstrate the effectiveness of our domain enhancement strategy:

\begin{itemize}[noitemsep,topsep=0pt]
\item \textbf{OCR Performance:} Stage 3 consistently improves OCR-related tasks, with particularly strong gains in handwritten text recognition (+8.20\%) and complex HTML table structure (+3.67\%). Even challenging tasks like structured information extraction show meaningful improvements (+2.68\%).

\item \textbf{Mathematical Reasoning:} Domain enhancement yields substantial improvements across all mathematical benchmarks. The most significant gains appear in our in-house datasets (+18.00\% and +7.18\%), while public benchmarks also benefit (Mathvision +6.02\%, Mathverse +2.61\%).

\item \textbf{Document Understanding:} Complex document tasks benefit significantly from domain-specific training. HTML table structure recognition improves by 3.40\% for simple tables and 3.67\% for complex tables, demonstrating enhanced structural understanding.

\item \textbf{Consistent Improvements:} All 16 evaluated tasks show positive gains with Stage 3, with no performance regressions observed. This validates that domain enhancement complements rather than conflicts with general training.
\end{itemize}

These results confirm that targeted domain enhancement in Stage 3 is crucial for achieving state-of-the-art performance in specialized tasks while maintaining general multimodal capabilities. The consistent improvements across diverse benchmarks validate our four-stage training strategy.

Importantly, Stage 3 training incurs relatively low computational costs while delivering significant performance improvements for domain-specific tasks.
This cost-effectiveness enables efficient customization for different application domains—organizations can start from our Stage 2 checkpoint and apply only Stage 3 and Stage 4 training with domain-specific data to achieve targeted enhancements.
This modular approach significantly reduces the barrier for developing specialized vision-language models tailored to specific industry needs.

\section{Limitations and Future Work}

While the Qianfan-VL series demonstrates strong performance across various benchmarks, several functional limitations remain that we plan to address in future iterations. The current models support a maximum context length of 32K tokens, which limits their ability to process lengthy documents, multi-page PDFs, or engage in extended multi-turn conversations. This constraint particularly affects applications requiring comprehensive document analysis or complex reasoning chains that span extensive textual and visual information. Additionally, despite various optimizations implemented throughout the training process, the models still require substantial computational resources for inference, especially when processing high-resolution images or multiple visual inputs simultaneously. This computational burden limits deployment scenarios in resource-constrained environments such as mobile devices or edge computing platforms. Furthermore, while the models excel at OCR and document understanding tasks, they currently lack certain advanced capabilities such as video understanding, 3D spatial reasoning, and fine-grained temporal analysis that are becoming increasingly important in real-world multimodal applications.

To address these limitations, we are pursuing several technical solutions and future research directions. We are actively working on extending the context window to 128K tokens and beyond through techniques like sparse attention mechanisms and hierarchical encoding strategies, which will enable the processing of entire books, lengthy technical documents, and maintenance of context across extended dialogues. For computational efficiency, we plan to integrate NaViT (Native Resolution ViT) techniques \citep{dehghani2024patch} to process images at their native resolutions without resizing, thereby reducing computational overhead while maintaining accuracy. We are also exploring quantization methods and model distillation approaches to create lighter variants suitable for edge deployment without significant performance degradation. In terms of capability expansion, future versions will incorporate specialized training for video understanding, 3D scene comprehension, and temporal reasoning. We also plan to enhance multilingual capabilities across more languages, and develop domain-specific variants for medical imaging, scientific diagrams, and technical blueprints through targeted fine-tuning strategies. These improvements will position the Qianfan-VL series as a more versatile and efficient solution for diverse multimodal understanding tasks across various domains and deployment scenarios.

\section{Conclusion}

Qianfan-VL represents a significant advancement in domain-enhanced vision-language models, successfully balancing general multimodal capabilities with specialized expertise in critical enterprise domains.
Through innovative training strategies, large-scale data synthesis, and efficient infrastructure utilization, we demonstrate that targeted domain enhancement can be achieved without sacrificing broad applicability.

The four-stage progressive training pipeline provides a principled approach to capability development, while our comprehensive data synthesis techniques ensure high-quality training data for specialized tasks.
The successful training on Kunlun chips validates the maturity of proprietary AI infrastructure for large-scale model development.

Qianfan-VL's strong performance on both general and domain-specific benchmarks, combined with flexible deployment options across different model sizes, makes it a practical solution for diverse enterprise multimodal applications.
Detailed application showcases demonstrating the model's capabilities in document processing, educational scenarios, and business intelligence are provided in Appendix A.
As we continue to expand capabilities and optimize performance, Qianfan-VL aims to bridge the gap between research advances and real-world deployment needs.

\section*{Acknowledgments}

We thank the Baidu AI Cloud team for infrastructure support, the Baige and Kunlun teams for AI infrastructure assistance, and all contributors to the QianFan platform. We are deeply grateful to the operations, storage, and network teams for maintaining the stability of the P800 clusters, enabling Qianfan-VL to train successfully at massive scale on 5000+ P800 chips. Special thanks to our annotation teams and quality assurance engineers for their meticulous work in data validation. There are many more colleagues who contributed to this project's success whom we cannot acknowledge individually.

\clearpage
\section*{Contributors}

\subsection*{Core Contributors}
\begin{multicols}{2}
\begin{itemize}[noitemsep,topsep=0pt,leftmargin=*]
\item Daxiang Dong$^*$
\item Mingming Zheng
\item Dong Xu
\item Bairong Zhuang
\item Wenyu Zhang
\item Chunhua Luo
\item Haoran Wang
\item Zijian Zhao
\item Jie Li
\item Yuxuan Li
\item Hanjun Zhong
\item Mengyue Liu
\item Jieting Chen
\item Shupeng Li
\item Jianmin Wu
\end{itemize}
\end{multicols}

\subsection*{Contributors}
\begin{multicols}{2}
\begin{itemize}[noitemsep,topsep=0pt,leftmargin=*]
\item Lun Tian
\item Yaping Feng
\item Xin Li
\item Donggang Jiang
\item Yong Chen
\item Yehua Xu
\item Duohao Qin
\item Chen Feng
\item Dan Wang
\item Henghua Zhang
\item Jingjing Ha
\item Jinhui He
\item Yanfeng Zhai
\item Chengxin Zheng
\item Jiayi Mao
\item Jiacheng Chen
\item Ruchang Yao
\item Ziye Yuan
\item Guangjun Xie$^{**}$
\item Dou Shen$^{**}$
\end{itemize}
\end{multicols}

\vspace{0.5em}
\noindent
$^*$ Project Lead\\
$^{**}$ Project Sponsor

\bibliography{qianfan_vl}

\clearpage
\appendix
\section{Application Showcases}

This appendix presents detailed examples showcasing Qianfan-VL's capabilities across diverse enterprise scenarios and real-world applications.

\subsection{Intelligent Document Processing}

\textbf{Scene Text Recognition:}

This example demonstrates Qianfan-VL's advanced OCR capabilities in challenging real-world scenarios:
\begin{itemize}[noitemsep,topsep=0pt]
\item \textbf{Model Capabilities:} Complex scene text detection, multi-directional text recognition, and handling of various fonts and sizes in natural images
\item \textbf{Technical Challenges:} The image contains text at different angles, varying lighting conditions, partial occlusions, and multilingual content with special characters
\item \textbf{Performance Highlights:} Accurate extraction of all visible text including phone numbers, distances, and location descriptions despite visual complexity
\end{itemize}

\begin{table}[!htbp]
\centering
\footnotesize
\begin{tabular}{|m{0.48\textwidth}|m{0.48\textwidth}|}
\hline
\begin{minipage}[c]{0.46\textwidth}
\centering
\textbf{Image:} \\[0.2em]
\includegraphics[width=\textwidth]{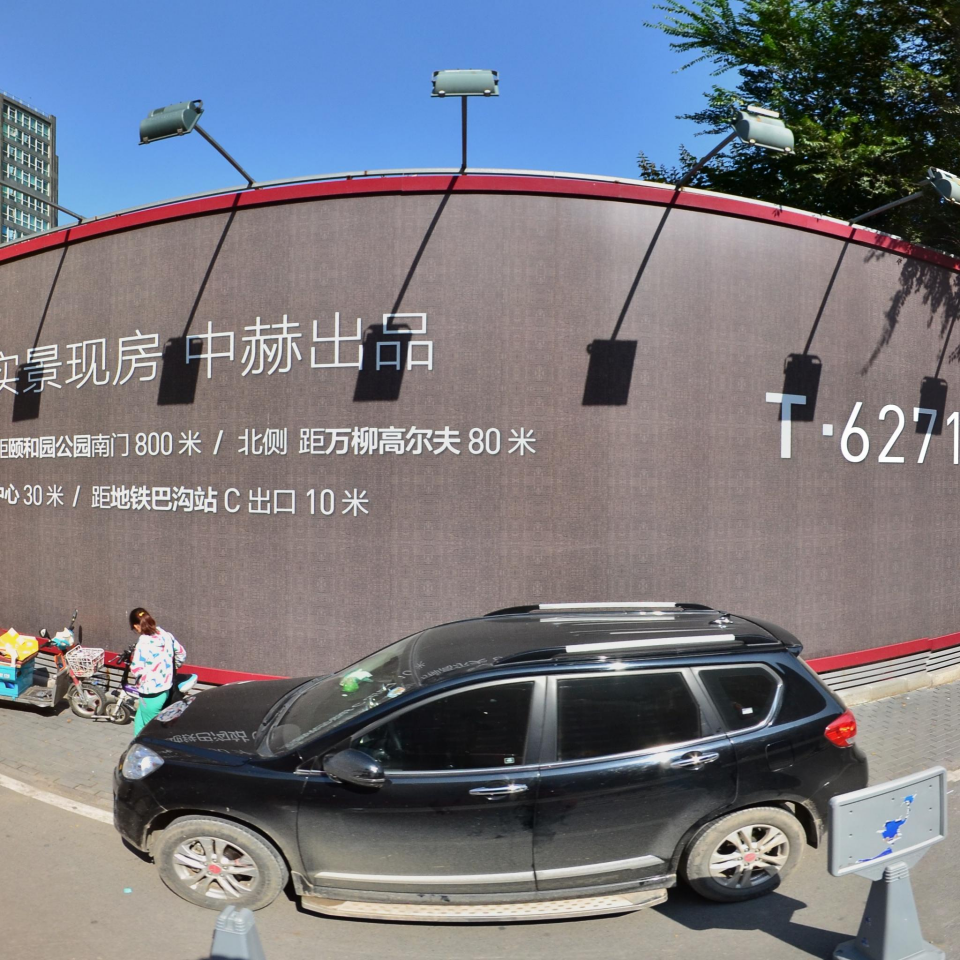}
\end{minipage} &
\begin{minipage}[c]{0.46\textwidth}
\begin{CJK}{UTF8}{gbsn}
\textbf{Prompt:} \\
请识别图中文字
\end{CJK}
\\[1em]
\begin{CJK}{UTF8}{gbsn}
\textbf{Response:} \\
实景现房中赫出品\\
颐和园公园南门800米/北侧距万柳高尔夫80米\\
心30米/距地铁巴沟站C出口10米\\
T·6271
\end{CJK}
\end{minipage} \\
\hline
\end{tabular}
\caption{Scene Text Recognition in Real Estate Advertisement}
\label{tab:scene_ocr}
\end{table}

\textbf{Invoice Information Extraction:}

This case showcases Qianfan-VL's structured information extraction and JSON generation capabilities:
\begin{itemize}[noitemsep,topsep=0pt]
\item \textbf{Model Capabilities:} Document layout understanding, table structure recognition, multi-language processing, and structured output generation
\item \textbf{Technical Challenges:} Complex invoice layouts with nested tables, multilingual text, numerical data extraction, and maintaining relationships between different invoice fields
\item \textbf{Performance Highlights:} Precise extraction of all invoice fields into well-formatted JSON, including item-level details with quantities and prices, while handling missing information gracefully
\end{itemize}

\begin{table}[H]
\centering
\footnotesize
\setlength{\tabcolsep}{3pt}
\begin{tabular}{|m{0.47\textwidth}|m{0.47\textwidth}|}
\hline
\begin{minipage}[c]{0.45\textwidth}
\centering
\textbf{Image:} \\[0.2em]
\includegraphics[width=0.95\textwidth]{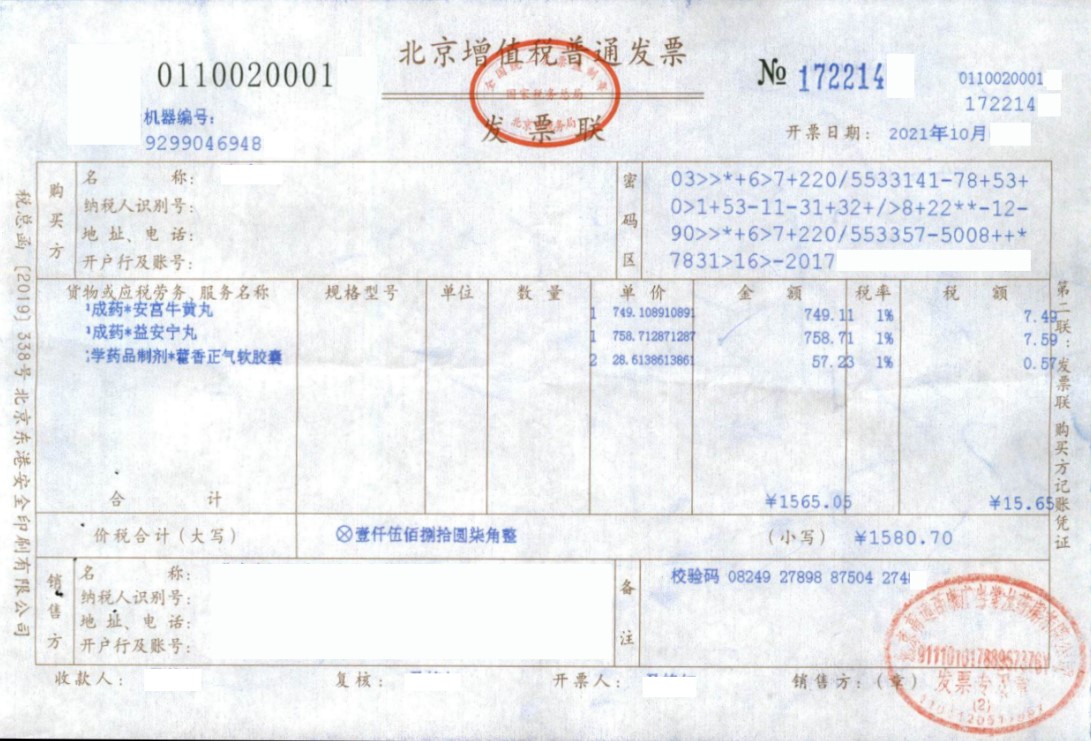}
\end{minipage} &
\begin{minipage}[c]{0.45\textwidth}
\begin{CJK}{UTF8}{gbsn}
\textbf{Prompt:} \\
你是发票提取专家。分析提供的图像（这是一个中文增值税发票），提取：\\
- 发票号码\\
- 开票日期 (YYYY-MM-DD)\\
- 供应商名称\\
- 购买方名称\\
- 总金额 (含货币)\\
- 项目列表：每个含描述、数量、单价、小计\\
缺失信息用"N/A"。仅输出JSON。
\end{CJK}
\\[0.5em]
\textbf{Response:} \\
\includegraphics[width=0.95\textwidth]{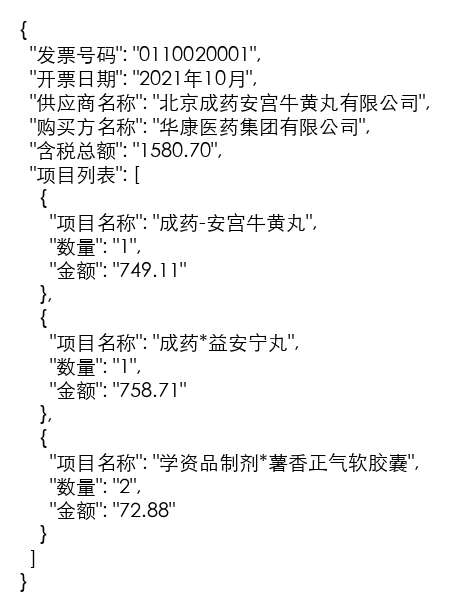}
\end{minipage} \\
\hline
\end{tabular}
\vspace{-0.5em}
\caption{Invoice Information Extraction with Structured JSON Output}
\label{tab:invoice_extraction}
\end{table}

\subsection{Educational Applications}

\textbf{Advanced Mathematical Reasoning:}

This example demonstrates Qianfan-VL's sophisticated mathematical understanding and problem-solving abilities:
\begin{itemize}[noitemsep,topsep=0pt]
\item \textbf{Model Capabilities:} Mathematical notation recognition, multi-step reasoning, geometric understanding, and step-by-step solution generation
\item \textbf{Technical Challenges:} Complex quadratic function analysis, coordinate system interpretation, multiple solution paths, and clear mathematical explanation in multiple languages
\item \textbf{Performance Highlights:} Complete solution with vertex formula application, discriminant analysis, and systematic verification of all answer choices with detailed reasoning
\end{itemize}

\begin{table}[!htbp]
\centering
\footnotesize
\begin{tabular}{|p{0.95\textwidth}|}
\hline
\textbf{Image:} \\
\begin{center}
\includegraphics[width=0.6\textwidth]{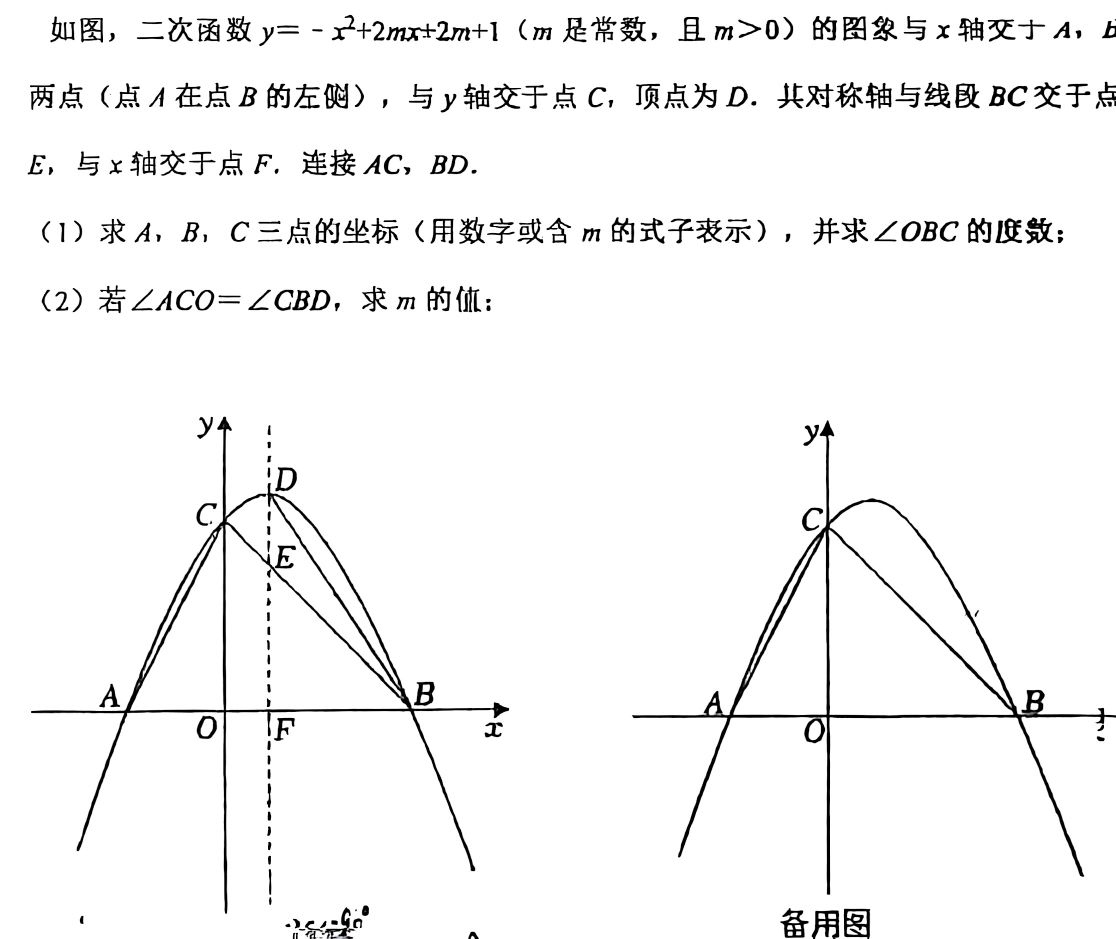}
\end{center} \\
\hline
\begin{CJK}{UTF8}{gbsn}
\textbf{Prompt:} 帮我求解这道题，给出详细步骤。
\end{CJK} \\
\hline
\textbf{Response:} \\
\begin{center}
\includegraphics[width=0.9\textwidth]{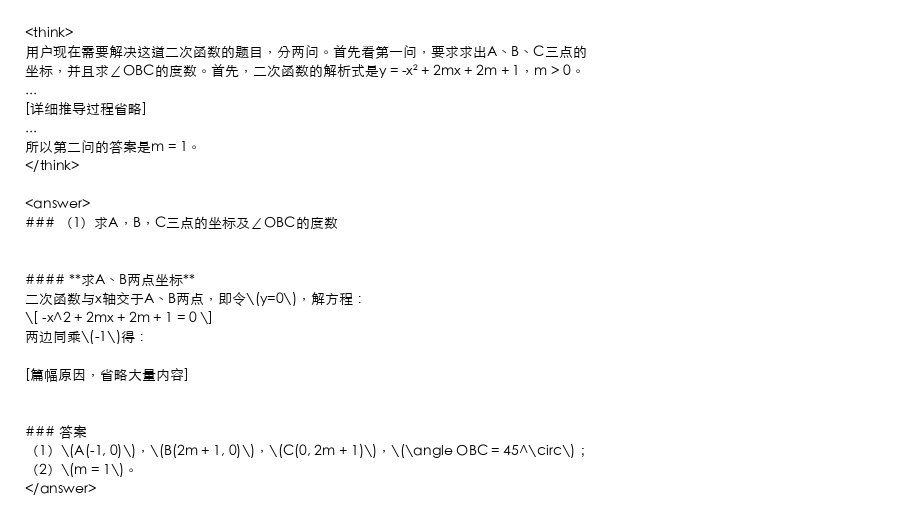}
\end{center} \\
\hline
\end{tabular}
\caption{Advanced Mathematical Reasoning}
\label{tab:math_reasoning}
\end{table}

\textbf{Technical Diagram Analysis:}

\begin{table}[H]
\centering
\footnotesize
\begin{tabular}{|m{0.48\textwidth}|m{0.48\textwidth}|}
\hline
\begin{minipage}[c]{0.46\textwidth}
\centering
\textbf{Image:} \\[0.2em]
\includegraphics[width=\textwidth]{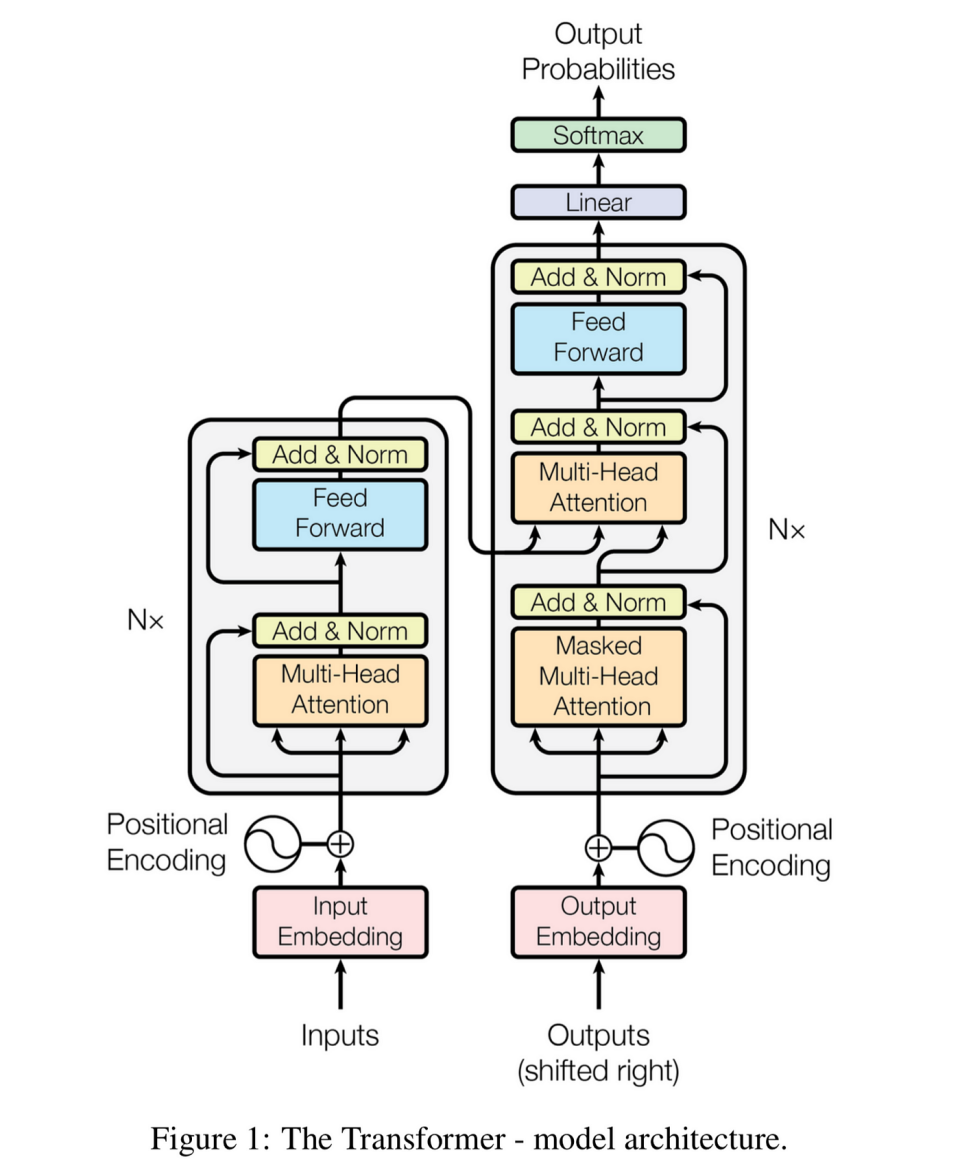}
\end{minipage} &
\begin{minipage}[c]{0.46\textwidth}
\begin{CJK}{UTF8}{gbsn}
\textbf{Prompt:} \\
非常详细解释这张图表的含义，解释清楚每一个元素。
\end{CJK}
\\[1em]
\begin{CJK}{UTF8}{gbsn}
\textbf{Response:} \\
这张图展示了Transformer模型的架构，这是一种广泛应用于自然语言处理和其他序列任务的深度学习模型。我将详细解释图中的每个元素：

\textbf{整体结构}\\
图1展示了Transformer模型的基本架构，包括编码器（左侧）和解码器（右侧）的组件。Transformer模型由多个这样的编码器-解码器模块堆叠而成（图中标注为"N×"，表示重复多次）。

\textbf{左侧：编码器（Encoder）}\\
1. \textbf{输入嵌入（Input Embedding）}：接收输入序列（如词序列），将每个词转换为固定维度的向量表示。\\
2. \textbf{位置编码（Positional Encoding）}：添加表示序列中元素位置的信息。\\
3. \textbf{多头注意力（Multi-Head Attention）}：允许模型关注输入序列的不同位置。\\
4. \textbf{前馈神经网络（Feed Forward）}：包含ReLU激活函数的简单神经网络。\\
5. \textbf{残差连接与归一化（Add \& Norm）}：残差连接和层归一化处理。

\textbf{右侧：解码器（Decoder）}\\
包括额外的掩码多头注意力和编码器-解码器注意力组件，用于序列生成任务。

\textbf{输出层}\\
线性层和Softmax层将解码器输出转换为概率分布。

核心创新在于自注意力机制，使模型能够高效处理长序列并捕捉不同位置间的关系。
\end{CJK}
\end{minipage} \\
\hline
\end{tabular}
\caption{Technical Architecture Diagram Analysis and Explanation}
\label{tab:transformer_analysis}
\end{table}

This case highlights Qianfan-VL's ability to understand and explain complex technical architectures:
\begin{itemize}[noitemsep,topsep=0pt]
\item \textbf{Model Capabilities:} Technical diagram interpretation, component relationship understanding, data flow analysis, and technical documentation generation
\item \textbf{Technical Challenges:} Understanding transformer architecture details, identifying component connections, explaining attention mechanisms, and providing implementation-ready descriptions
\item \textbf{Performance Highlights:} Accurate identification of all architectural components, clear explanation of information flow, and practical code structure suggestions
\end{itemize}

\textbf{Algorithm Theory Explanation:}

\begin{table}[H]
\centering
\footnotesize
\begin{tabular}{|m{0.48\textwidth}|m{0.48\textwidth}|}
\hline
\begin{minipage}[c]{0.46\textwidth}
\centering
\textbf{Images:} \\[0.2em]
\includegraphics[width=\textwidth]{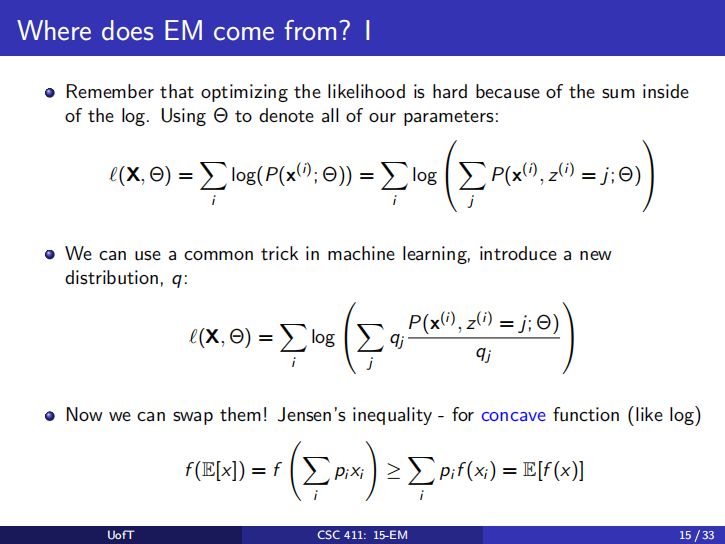}
\\[0.5em]
\includegraphics[width=\textwidth]{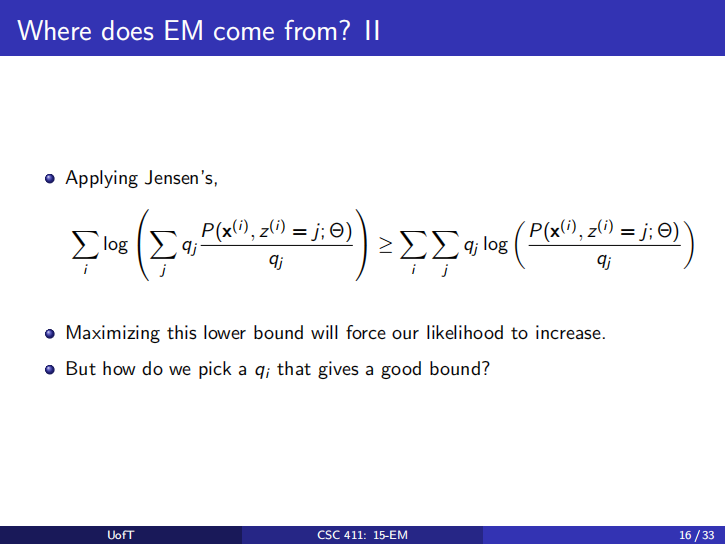}
\end{minipage} &
\begin{minipage}[c]{0.46\textwidth}
\begin{CJK}{UTF8}{gbsn}
\textbf{Prompt:} \\
帮我解释下面这些slides的详细内容。\texttt{\textless think\textgreater}
\end{CJK}
\\[1em]
\textbf{Response:} \\
\includegraphics[width=\textwidth]{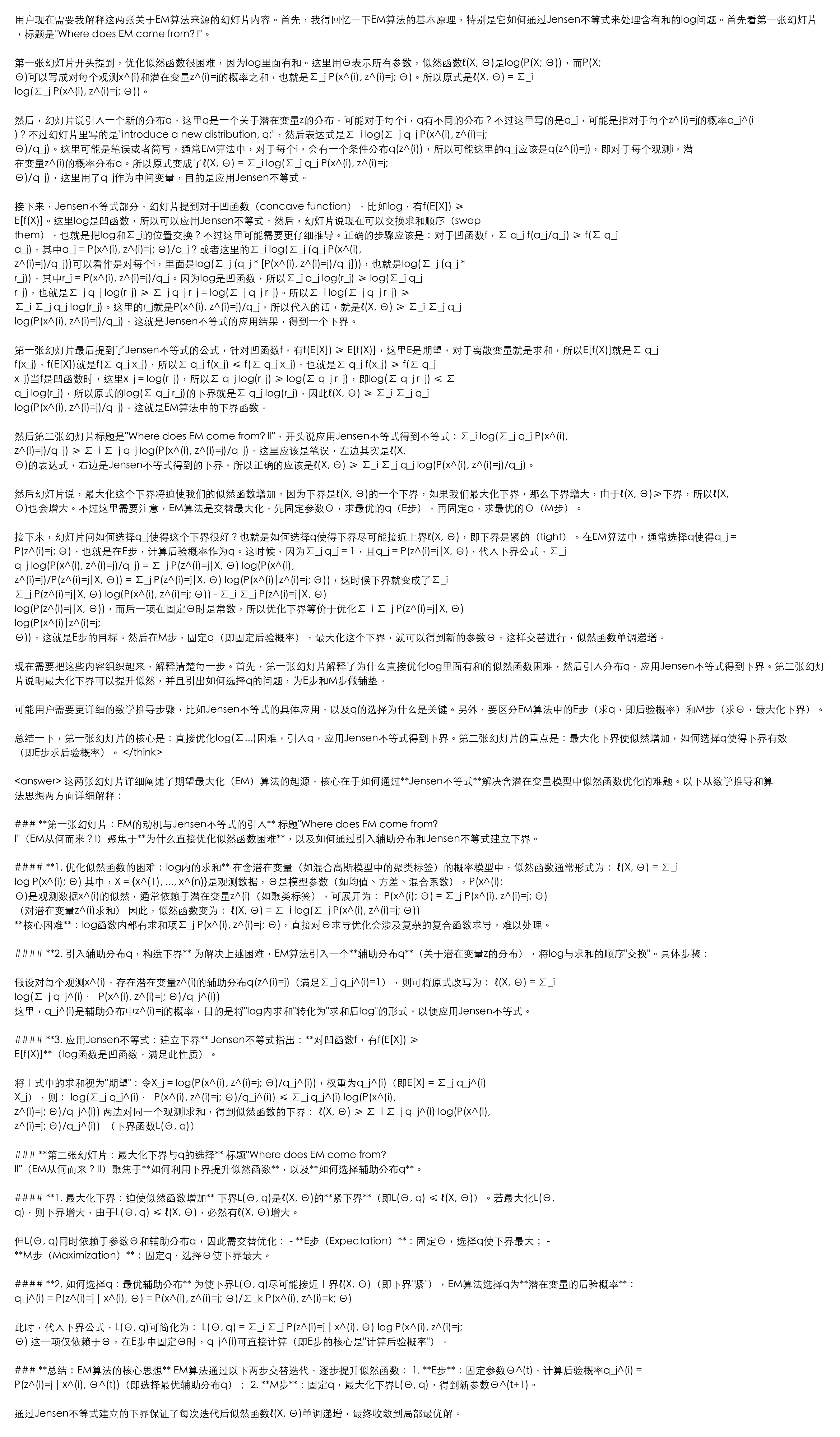}
\end{minipage} \\
\hline
\end{tabular}
\caption{Algorithm Theory Analysis and Mathematical Derivation Explanation}
\label{tab:algorithm_theory}
\end{table}

This example showcases Qianfan-VL's deep understanding of machine learning algorithms and mathematical concepts:
\begin{itemize}[noitemsep,topsep=0pt]
\item \textbf{Model Capabilities:} Algorithm comprehension, mathematical formula interpretation, multi-slide correlation, and theoretical explanation with Chain-of-Thought reasoning
\item \textbf{Technical Challenges:} Understanding EM algorithm mathematics, connecting concepts across multiple slides, explaining complex iterative processes, and providing intuitive interpretations
\item \textbf{Performance Highlights:} Comprehensive explanation of E-step and M-step, clear mathematical derivations, and practical insights into convergence properties and applications
\end{itemize}

\subsection{Business Intelligence}

\textbf{Political Polling Analysis:}

This case demonstrates Qianfan-VL's advanced data visualization interpretation and trend analysis capabilities:
\begin{itemize}[noitemsep,topsep=0pt]
\item \textbf{Model Capabilities:} Time-series data interpretation, multi-variable trend analysis, contextual understanding of political events, and comprehensive data storytelling
\item \textbf{Technical Challenges:} Identifying multiple data series, correlating trends with historical events, extracting precise percentage values, and providing meaningful interpretations
\item \textbf{Performance Highlights:} Accurate identification of all political parties, precise trend description with timeline markers, and insightful analysis of electoral implications
\end{itemize}

\begin{table}[!htbp]
\centering
\footnotesize
\begin{tabular}{|m{0.48\textwidth}|m{0.48\textwidth}|}
\hline
\begin{minipage}[c]{0.46\textwidth}
\centering
\textbf{Image:} \\[0.2em]
\includegraphics[width=\textwidth]{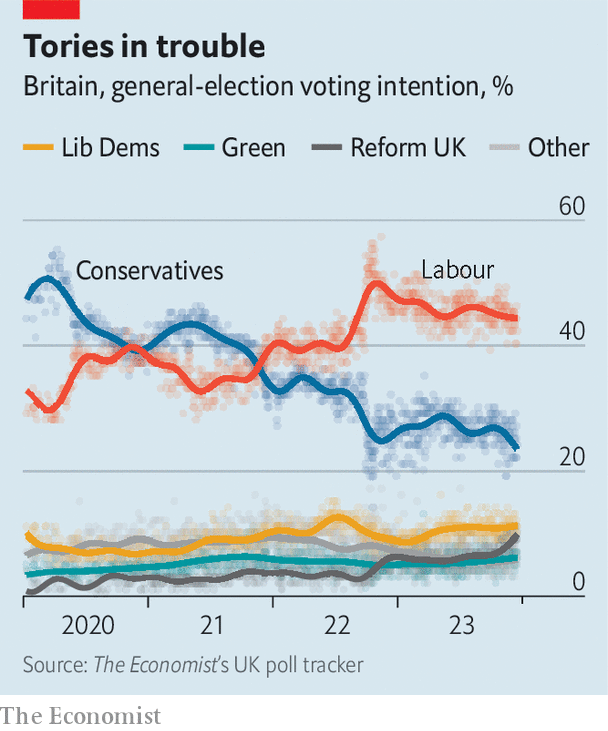}
\end{minipage} &
\begin{minipage}[c]{0.46\textwidth}
\begin{CJK}{UTF8}{gbsn}
\textbf{Prompt:} \\
帮我详细解释这张图表的含义。
\end{CJK}
\\[1em]
\begin{CJK}{UTF8}{gbsn}
\textbf{Response:} \\
这张图表展示了英国主要政党的选举投票意向变化趋势，数据来源于经济学人杂志的英国民意调查追踪器。图表标题为"困境中的保守党"（Tories in trouble），副标题说明这是英国大选投票意向百分比。

图表主要内容包括：

1. 两条主要曲线：
   - 红色曲线代表工党（Labour）支持率
   - 蓝色曲线代表保守党（Conservatives）支持率

2. 下方还有四条较小的曲线，分别代表：
   - 黄色：自由民主党（Lib Dems）
   - 青色：绿党（Green）
   - 灰色：改革英国党（Reform UK）
   - 浅灰色：其他政党（Other）

3. 横轴表示时间，从2020年到2023年
4. 纵轴表示支持率百分比，从0\%到60\%

从图表可以观察到：
- 保守党（蓝色）在2020年初支持率较高（约45-50\%），但呈现明显下降趋势，到2023年降至约25\%左右
- 工党（红色）在2020年时支持率较低（约30-35\%），但整体呈上升趋势，到2023年上升至约45\%左右
- 两党支持率大约在2021-2022年间出现交叉，之后工党保持领先
- 其他小党的支持率相对稳定，基本维持在较低水平

总体趋势表明，保守党的支持率在下降，而工党的支持率在上升，标题"困境中的保守党"也反映了这一政治动态。
\end{CJK}
\end{minipage} \\
\hline
\end{tabular}
\caption{Political Polling Data Analysis and Trend Interpretation}
\label{tab:political_polling}
\end{table}

This example demonstrates Qianfan-VL's sophisticated capability to analyze complex multi-party political polling charts:
\begin{itemize}[noitemsep,topsep=0pt]
\item \textbf{Multi-trend analysis:} Simultaneous tracking and interpretation of multiple political parties' polling trends over time
\item \textbf{Cross-referencing data sources:} Recognition and attribution of data sources (The Economist's UK poll tracker)
\item \textbf{Temporal pattern recognition:} Identification of trend reversals, crossover points, and long-term trajectories across different time periods
\item \textbf{Contextual interpretation:} Understanding of political implications and providing meaningful analysis beyond raw data extraction
\end{itemize}

\textbf{Geographical Data Analysis:}

This example demonstrates Qianfan-VL's capability to interpret complex geographical and meteorological visualizations:
\begin{itemize}[noitemsep,topsep=0pt]
\item \textbf{Model Capabilities:} Heat map interpretation, geographical feature recognition, temperature scale understanding, and regional climate analysis
\item \textbf{Technical Challenges:} Color gradient mapping to temperature values, identifying geographical boundaries, understanding seasonal patterns, and providing location-specific insights
\item \textbf{Performance Highlights:} Accurate temperature range identification, detailed regional analysis, and comprehensive explanation of climate distribution patterns
\end{itemize}

\begin{table}[H]
\centering
\footnotesize
\begin{tabular}{|m{0.48\textwidth}|m{0.48\textwidth}|}
\hline
\begin{minipage}[c]{0.46\textwidth}
\centering
\textbf{Image:} \\[0.2em]
\includegraphics[width=\textwidth]{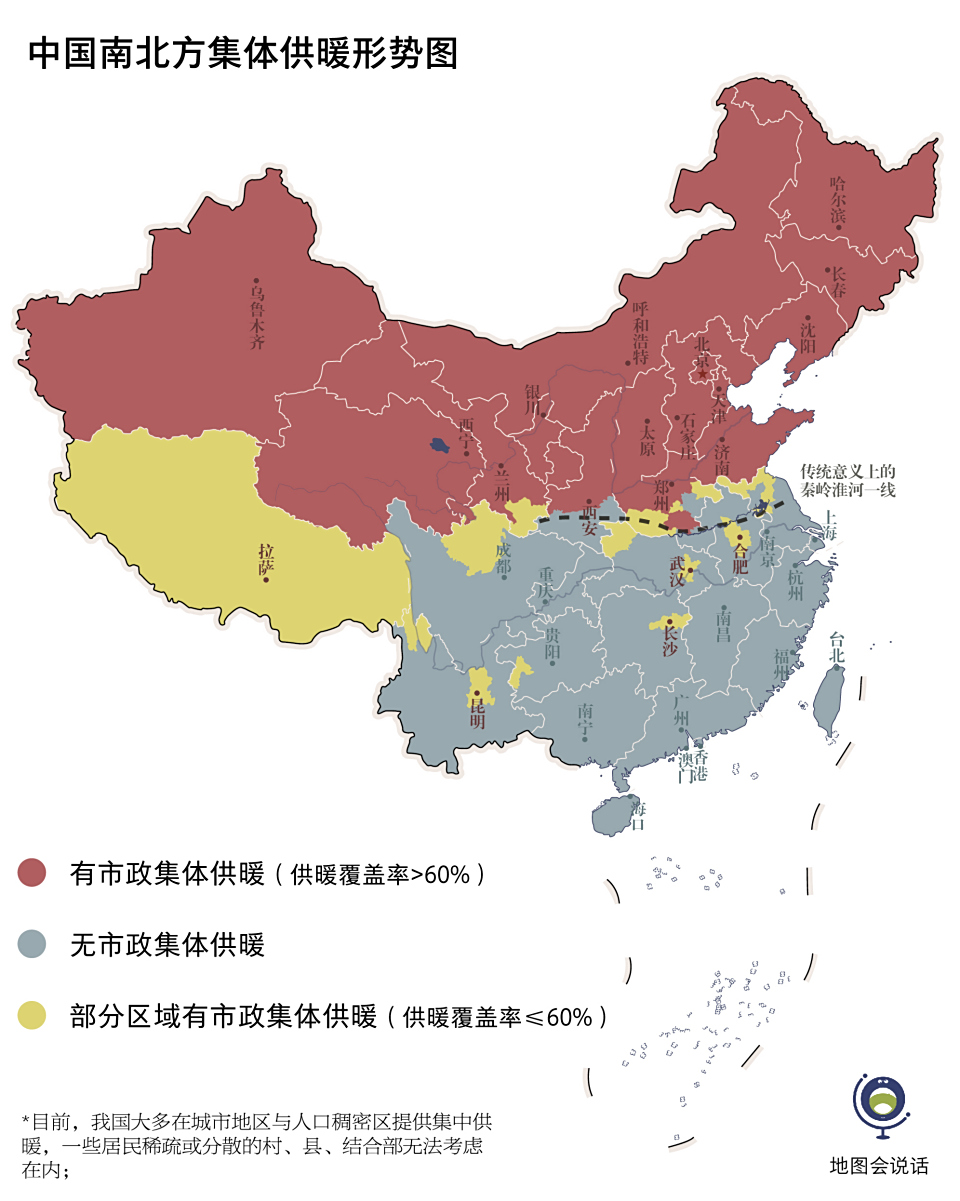}
\end{minipage} &
\begin{minipage}[c]{0.46\textwidth}
\begin{CJK}{UTF8}{gbsn}
\textbf{Prompt:} \\
云南有任何区域有集体供暖吗？\texttt{\textless think\textgreater}
\end{CJK}
\\[1em]
\begin{CJK}{UTF8}{gbsn}
\textbf{Response:} \\
用户问的是"云南有任何区域有集体供暖吗？"首先，我需要查看提供的地图信息。地图的标题是"中国南北方集体供暖形势图"，图例分为三种：红色是有市政集体供暖（覆盖率>60\%），灰色是无市政集体供暖，黄色是部分区域有市政集体供暖（覆盖率≤60\%）。

首先，找到云南省的位置。云南省位于中国西南，地图上昆明是云南省的省会。在地图上，昆明所在的区域颜色需要确认。根据图例，黄色代表部分区域有市政集体供暖（覆盖率≤60\%）。查看地图，昆明被标记为黄色，而云南省的其他区域可能大部分是灰色（无市政集体供暖），但问题问的是"任何区域"，只要有一个区域就算。

另外，地图下方的说明提到"目前，我国大多在城市地区与人口稠密区提供集中供暖"，昆明作为云南省的省会，属于较大城市，可能属于部分区域有集体供暖的情况，因为昆明的颜色是黄色。所以结论应该是云南的部分区域（如昆明）有市政集体供暖，属于黄色区域，即覆盖率≤60\%。需要确认地图上昆明是否确实是黄色，从描述中"昆明"被标记为黄色，所以答案应该是肯定的，云南有部分区域（如昆明）有市政集体供暖。
\texttt{\textless/think\textgreater}

\texttt{\textless answer\textgreater}
根据"中国南北方集体供暖形势图"及图例信息：

云南省的省会\textbf{昆明}所在区域在地图上被标记为\textbf{黄色}，对应图例中"部分区域有市政集体供暖（供暖覆盖率≤60\%）"。

因此，云南的部分区域（如昆明）存在市政集体供暖，属于"部分区域有集体供暖"的范畴。

\textbf{答案：云南的部分区域（如昆明）有市政集体供暖（供暖覆盖率≤60\%）。}
\texttt{\textless/answer\textgreater}
\end{CJK}
\end{minipage} \\
\hline
\end{tabular}
\caption{Geographical Map Analysis and Regional Information Extraction}
\label{tab:geographical_analysis}
\end{table}

This example showcases Qianfan-VL's capability to analyze geographical maps and extract regional information:
\begin{itemize}[noitemsep,topsep=0pt]
\item \textbf{Map legend interpretation:} Accurate understanding of color-coded legends and their corresponding meanings in geographical context
\item \textbf{Spatial reasoning:} Identification of specific regions (Yunnan province, Kunming city) and their geographical relationships within the map
\item \textbf{Data cross-referencing:} Matching visual map elements with legend information to provide accurate regional analysis
\item \textbf{Localized information extraction:} Providing specific, targeted answers about regional characteristics based on map data
\end{itemize}

\textbf{Advertisement Analysis:}

This case showcases Qianfan-VL's understanding of marketing materials and brand communication strategies:
\begin{itemize}[noitemsep,topsep=0pt]
\item \textbf{Model Capabilities:} Advertisement layout analysis, brand messaging interpretation, visual hierarchy understanding, and marketing strategy identification
\item \textbf{Technical Challenges:} Analyzing multi-element compositions, understanding cultural context, identifying target audience messaging, and interpreting visual metaphors
\item \textbf{Performance Highlights:} Complete identification of all visual elements, understanding of brand positioning, and insightful analysis of marketing effectiveness
\end{itemize}

\begin{table}[H]
\centering
\footnotesize
\begin{tabular}{|m{0.48\textwidth}|m{0.48\textwidth}|}
\hline
\begin{minipage}[c]{0.46\textwidth}
\centering
\textbf{Image:} \\[0.2em]
\includegraphics[width=\textwidth]{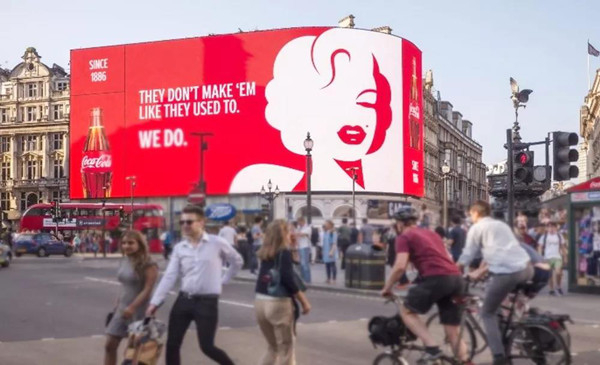}
\end{minipage} &
\begin{minipage}[c]{0.46\textwidth}
\begin{CJK}{UTF8}{gbsn}
\textbf{Prompt:} \\
请描述这张广告牌上的主要内容，并且分析一下这句广告词的含义。
\end{CJK}
\\[1em]
\begin{CJK}{UTF8}{gbsn}
\textbf{Response:} \\
广告牌上是可口可乐的广告，主色调为红色，上面有一个简约风格的女性头像剪影，广告词是'THEY DON'T MAKE 'EM LIKE THEY USED TO. WE DO.'（他们不再制造像过去那样的产品了。但我们仍然制造。）

这句广告词的含义是强调可口可乐始终保持着其原始配方和经典口味，暗示其他饮料品牌可能已经改变了原有配方或品质，但可口可乐依然坚持传统，保持着最初的味道和品质。广告通过怀旧情感唤起消费者对经典产品的认同和信任。
\end{CJK}
\end{minipage} \\
\hline
\end{tabular}
\caption{Advertisement Content Analysis and Marketing Message Interpretation}
\label{tab:advertisement_analysis}
\end{table}

This example demonstrates Qianfan-VL's sophisticated capability to analyze advertising content and interpret marketing strategies:
\begin{itemize}[noitemsep,topsep=0pt]
\item \textbf{Visual brand recognition:} Accurate identification of brand elements, color schemes, and design aesthetics in complex urban advertising displays
\item \textbf{Text extraction and translation:} Precise reading of advertising copy and providing contextual translation for cross-linguistic understanding
\item \textbf{Marketing strategy analysis:} Deep interpretation of advertising messages, understanding the psychological appeal and competitive positioning
\item \textbf{Cultural and emotional context:} Recognition of nostalgic marketing approaches and their intended impact on consumer sentiment and brand loyalty
\end{itemize}

\textbf{Media Information Extraction:}

This example illustrates Qianfan-VL's ability to extract and analyze information from media promotional materials:
\begin{itemize}[noitemsep,topsep=0pt]
\item \textbf{Model Capabilities:} Media content understanding, celebrity recognition, program information extraction, and entertainment industry knowledge
\item \textbf{Technical Challenges:} Processing stylized text layouts, understanding program formats, extracting scheduling information, and identifying cast members
\item \textbf{Performance Highlights:} Accurate extraction of all program details, understanding of show format and genre, and comprehensive cast identification
\end{itemize}

\begin{table}[H]
\centering
\footnotesize
\begin{tabular}{|m{0.48\textwidth}|m{0.48\textwidth}|}
\hline
\begin{minipage}[c]{0.46\textwidth}
\centering
\textbf{Image:} \\[0.2em]
\includegraphics[width=\textwidth]{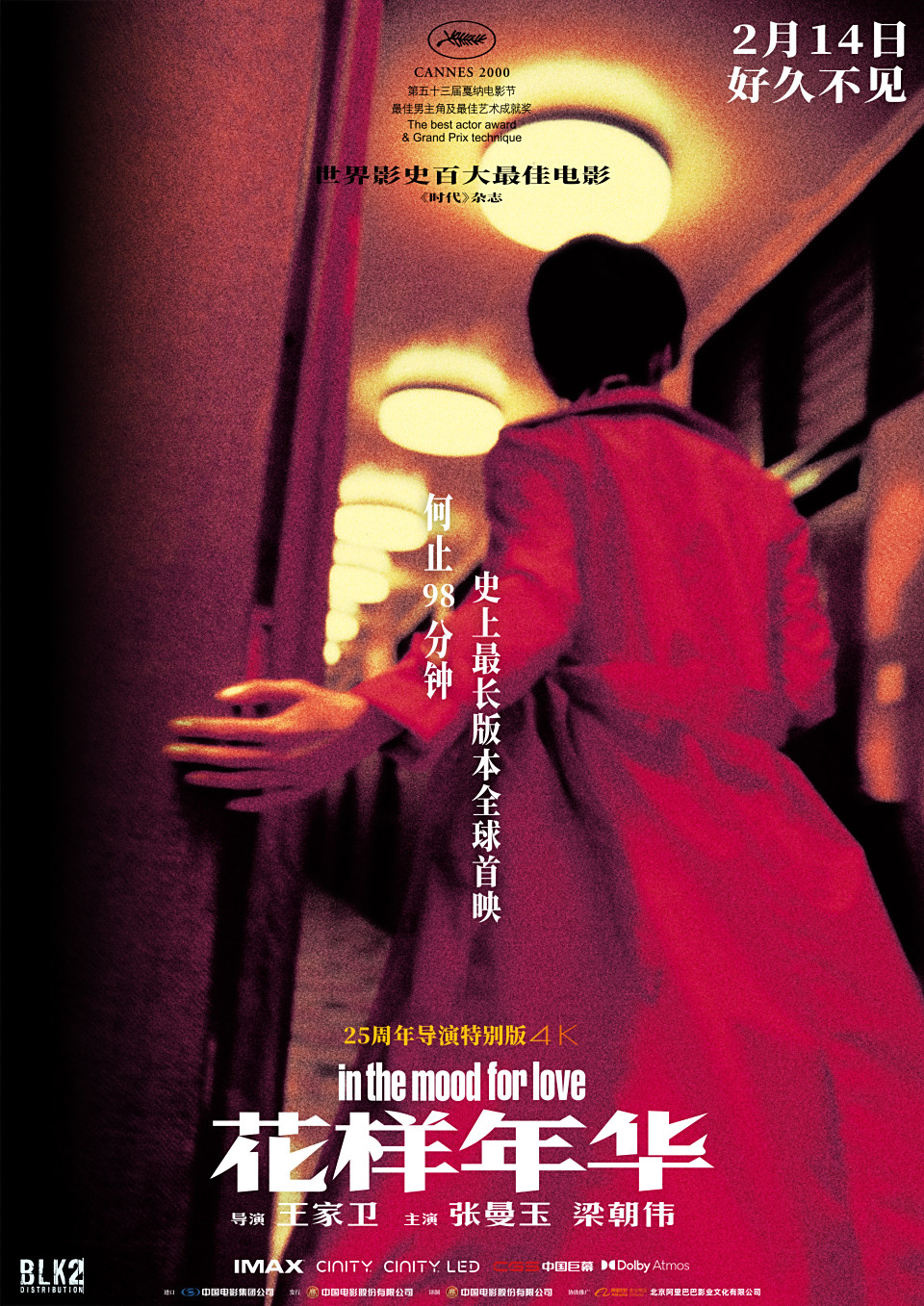}
\end{minipage} &
\begin{minipage}[c]{0.46\textwidth}
\begin{CJK}{UTF8}{gbsn}
\textbf{Prompt:} \\
请抽取这张电影海报中所有的关键信息。
\end{CJK}
\\[1em]
\begin{CJK}{UTF8}{gbsn}
\textbf{Response:} \\
这张电影海报是王家卫导演的《花样年华》(In the Mood for Love)的宣传海报。以下是我从海报中抽取的关键信息：

1. 电影名称：
   - 英文：In the Mood for Love
   - 中文：花样年华

2. 导演与主演：
   - 导演：王家卫
   - 主演：张曼玉、梁朝伟

3. 上映信息：
   - 上映日期：2月14日
   - 标语："好久不见"

4. 版本信息：
   - "25周年导演特别版4K"

5. 技术规格：
   - IMAX
   - CINITY
   - CINITY LED
   - 中国巨幕
   - Dolby Atmos

6. 电影节荣誉：
   - "CANNES 2000"（戛纳2000）
   - "最佳男主角及最佳艺术成就奖"（The best actor award \& Grand Prix technique）
   - "世界影史百大最佳电影"（《时代》杂志评选）

7. 其他文字信息：
   - "何止98分钟"
   - "史上最长版本全球首映"

海报整体以红色为主色调，展现了一个背对观众的人物形象，站在走廊中，符合王家卫电影特有的美学风格和氛围。
\end{CJK}
\end{minipage} \\
\hline
\end{tabular}
\caption{Movie Poster Information Extraction and Content Analysis}
\label{tab:movie_poster}
\end{table}

This example showcases Qianfan-VL's comprehensive capability to extract and analyze multimedia content information:
\begin{itemize}[noitemsep,topsep=0pt]
\item \textbf{Multilingual text recognition:} Accurate extraction of multilingual text from complex poster layouts with varying fonts and sizes
\item \textbf{Structured information organization:} Systematic categorization of extracted information including titles, credits, release details, and technical specifications
\item \textbf{Cultural and artistic analysis:} Recognition of directorial style, aesthetic elements, and cultural significance of the film poster
\item \textbf{Complete metadata extraction:} Comprehensive identification of all relevant information including awards, technical formats, and promotional elements
\end{itemize}

\textbf{Financial Policy Analysis:}

This case demonstrates Qianfan-VL's ability to understand and analyze complex financial policy presentations:
\begin{itemize}[noitemsep,topsep=0pt]
\item \textbf{Model Capabilities:} Financial terminology understanding, policy interpretation, structured information extraction, and professional document analysis
\item \textbf{Technical Challenges:} Processing dense financial content, understanding regulatory language, extracting key policy points, and maintaining logical structure
\item \textbf{Performance Highlights:} Accurate extraction of all policy details, clear categorization of financial measures, and comprehensive understanding of implementation guidelines
\end{itemize}

\begin{table}[H]
\centering
\footnotesize
\begin{tabular}{|m{0.48\textwidth}|m{0.48\textwidth}|}
\hline
\begin{minipage}[c]{0.46\textwidth}
\centering
\textbf{Image:} \\[0.2em]
\includegraphics[width=\textwidth]{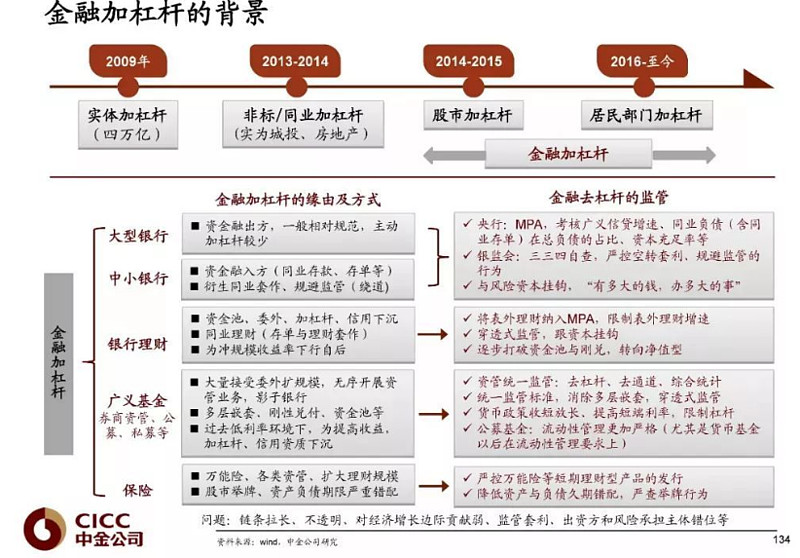}
\end{minipage} &
\begin{minipage}[c]{0.46\textwidth}
\begin{CJK}{UTF8}{gbsn}
\textbf{Prompt:} \\
请分条回答：金融去杠杆的监管是如何应对广义基金的金融加杠杆的？
\end{CJK}
\\[1em]
\begin{CJK}{UTF8}{gbsn}
\textbf{Response:} \\
1. 资管统一监管：去杠杆、去通道、综合统计

2. 统一监管标准：消除多层嵌套，穿透式监管

3. 货币政策收紧：提高市场利率，限制杠杆

4. 公募基金：流动性管理更加严格（尤其是货币基金以后在流动性管理要求上）
\end{CJK}
\end{minipage} \\
\hline
\end{tabular}
\caption{Financial Policy and Regulatory Analysis}
\label{tab:financial_policy}
\end{table}

This example demonstrates Qianfan-VL's expertise in analyzing complex financial and regulatory policy documents:
\begin{itemize}[noitemsep,topsep=0pt]
\item \textbf{Policy comprehension:} Deep understanding of financial regulatory frameworks including deleveraging mechanisms and asset management oversight
\item \textbf{Structured policy analysis:} Systematic breakdown of regulatory approaches with clear categorization of different policy instruments
\item \textbf{Financial terminology mastery:} Accurate usage and explanation of specialized financial terms and asset management regulations including penetrating supervision mechanisms
\item \textbf{Regulatory impact assessment:} Understanding of how different policy measures address specific financial risks and leverage issues in the fund industry
\end{itemize}

\subsection{Structured Information Extraction}

\textbf{Table Structure Analysis and Data Extraction:}

This sophisticated example demonstrates Qianfan-VL's advanced capability to handle complex multi-table extraction tasks:
\begin{itemize}[noitemsep,topsep=0pt]
\item \textbf{Model Capabilities:} Complex table structure recognition, multi-table relationship understanding, structured JSON generation, and data validation
\item \textbf{Technical Challenges:} Processing multiple interconnected tables, maintaining data relationships across different measurement systems, handling dense numerical data, and ensuring output accuracy
\item \textbf{Performance Highlights:} Perfect extraction of all size conversions, accurate mapping between different sizing systems, and well-formatted JSON output with complete data integrity
\end{itemize}

\begin{table}[H]
\centering
\footnotesize
\begin{tabular}{|m{0.48\textwidth}|m{0.48\textwidth}|}
\hline
\begin{minipage}[c]{0.46\textwidth}
\centering
\textbf{Image:} \\[0.2em]
\includegraphics[width=\textwidth]{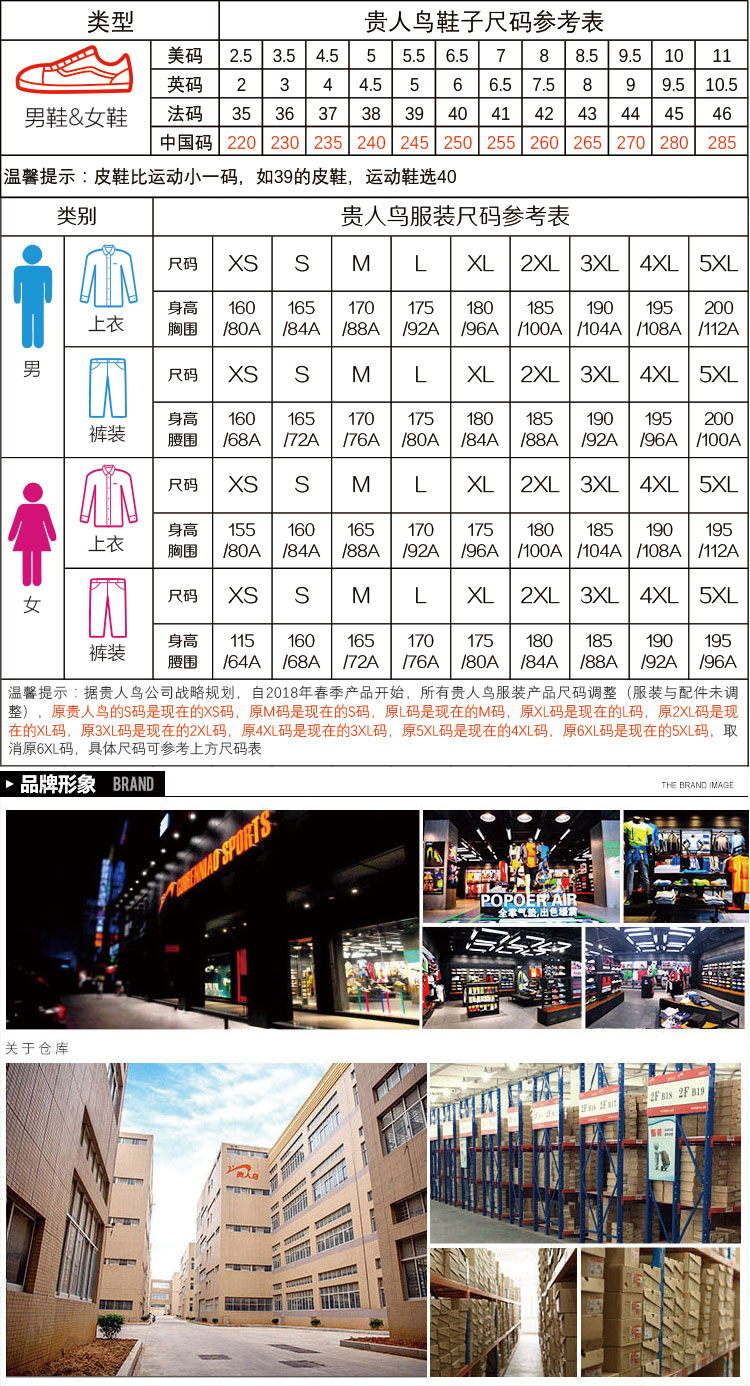}
\end{minipage} &
\begin{minipage}[t]{0.46\textwidth}
\begin{CJK}{UTF8}{gbsn}
\textbf{Prompt:} \\
请从图片中的"品牌形象"以上的图片部分，抽取以下关键信息：鞋子尺码对照表（包括美码、英码、法码、中国码的对应关系），服装尺码对照表（包括男女上衣和裤装的尺码、身高、胸围、腰围数据）。以JSON格式输出并且只输出JSON信息，不要输出其他内容，确保信息的准确性和完整性。
\end{CJK}
\\[1em]
\textbf{Response:} \\
\includegraphics[width=\textwidth,height=8cm,keepaspectratio]{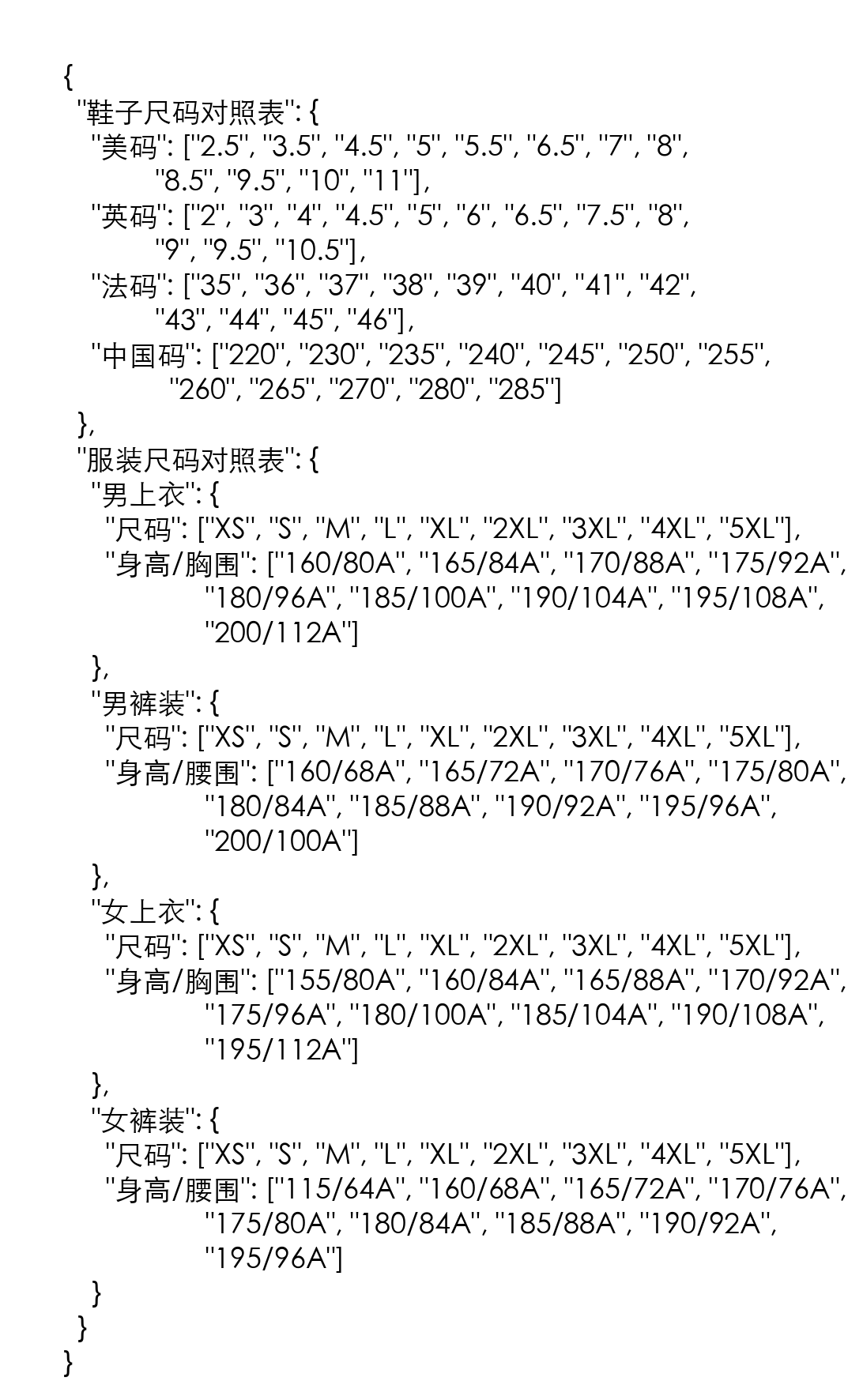}
\end{minipage} \\
\hline
\end{tabular}
\caption{Structured Information Extraction from Complex Tables}
\label{tab:structured_extraction}
\end{table}

This example demonstrates Qianfan-VL's capability to extract structured information from complex multi-table layouts:
\begin{itemize}[noitemsep,topsep=0pt]
\item \textbf{Multi-table processing:} Accurately identifies and processes multiple interconnected tables within a single image
\item \textbf{Cross-reference understanding:} Maintains relationships between different sizing systems and measurements
\item \textbf{Structured output generation:} Produces well-formatted JSON output with proper hierarchical organization
\item \textbf{Data integrity preservation:} Ensures accuracy of numerical data and dimensional correspondences across different measurement systems
\end{itemize}

\subsection{Sports Schedule Analysis}

\textbf{Event Calendar Information Extraction:}

This final example showcases Qianfan-VL's ability to extract and analyze complex scheduling information from sports calendars:
\begin{itemize}[noitemsep,topsep=0pt]
\item \textbf{Model Capabilities:} Calendar layout understanding, event scheduling interpretation, multi-competition tracking, and temporal relationship analysis
\item \textbf{Technical Challenges:} Processing dense calendar layouts, distinguishing between different competition types, extracting dates and venues, and understanding tournament structures
\item \textbf{Performance Highlights:} Complete extraction of all match schedules, accurate identification of competitions and venues, and clear temporal organization of events
\end{itemize}

\begin{table}[H]
\centering
\footnotesize
\begin{tabular}{|m{0.48\textwidth}|m{0.48\textwidth}|}
\hline
\begin{minipage}[c]{0.46\textwidth}
\centering
\textbf{Image:} \\[0.2em]
\includegraphics[width=\textwidth]{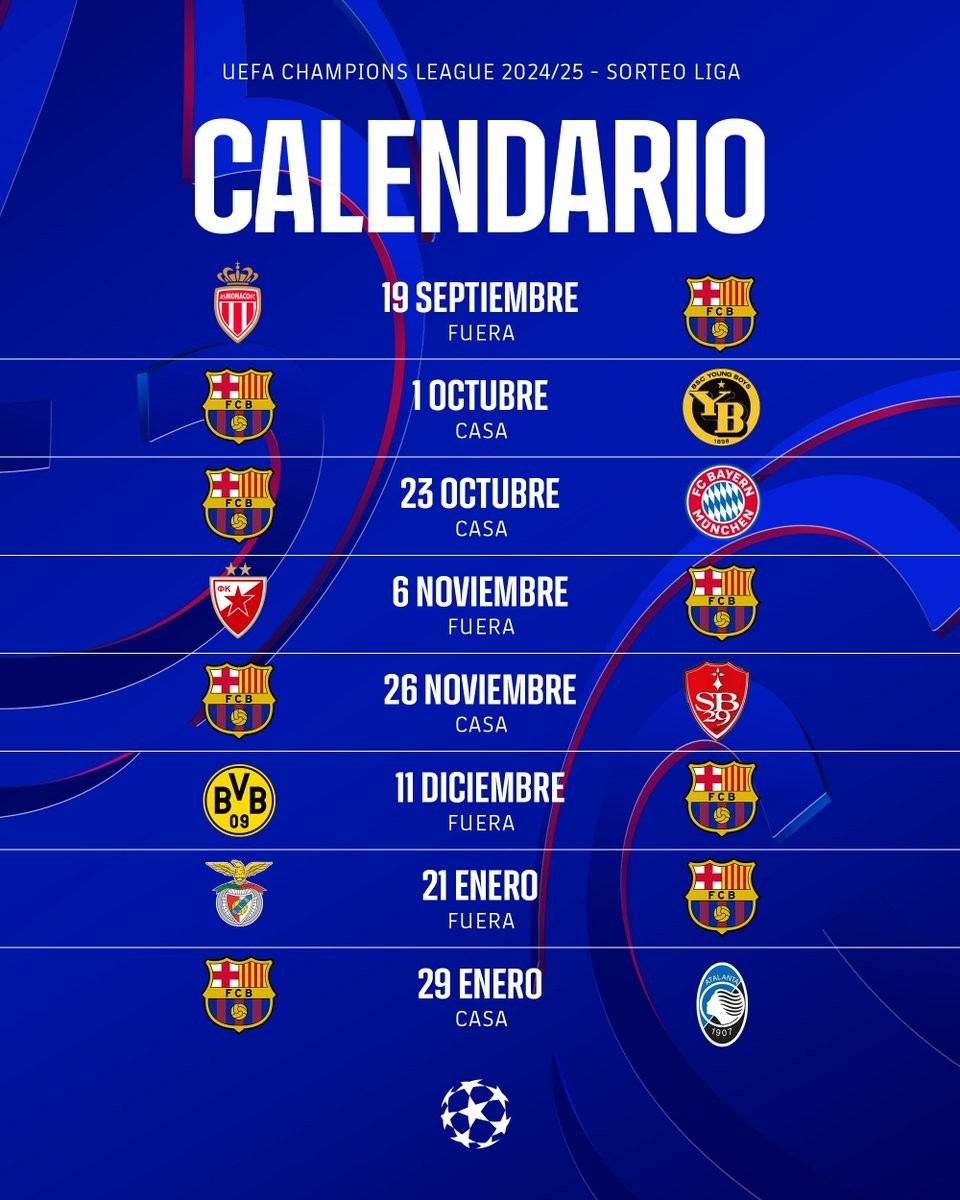}
\end{minipage} &
\begin{minipage}[c]{0.46\textwidth}
\begin{CJK}{UTF8}{gbsn}
\textbf{Prompt:} \\
巴塞罗那会在什么时候和德甲球队进行比赛？
\end{CJK}
\\[1em]
\begin{CJK}{UTF8}{gbsn}
\textbf{Response:} \\
10月23日（对阵拜仁慕尼黑）和12月11日（对阵多特蒙德）。
\end{CJK}
\end{minipage} \\
\hline
\end{tabular}
\caption{Sports Schedule Analysis and Information Extraction}
\label{tab:sports_schedule}
\end{table}

This example showcases Qianfan-VL's ability to analyze sports calendars and extract specific scheduling information:
\begin{itemize}[noitemsep,topsep=0pt]
\item \textbf{Calendar interpretation:} Accurately reads and interprets complex sports scheduling calendars with multiple teams and dates
\item \textbf{Selective information extraction:} Identifies specific matches based on league criteria (German teams in this case)
\item \textbf{Date and opponent recognition:} Precisely extracts match dates and opponent team names from visual calendar data
\item \textbf{Context-aware filtering:} Understands the relationship between team leagues and provides relevant match information
\end{itemize}

\end{document}